\documentclass{article}

\PassOptionsToPackage{numbers,sort&compress}{natbib}
\usepackage[eandd, final]{neurips_2026}

\usepackage[utf8]{inputenc} 
\usepackage[T1]{fontenc}    
\usepackage{hyperref}       
\usepackage{url}            
\usepackage{booktabs}       
\usepackage{amsfonts}       
\usepackage{amsmath,amssymb}
\usepackage{nicefrac}       
\usepackage{microtype}      
\usepackage{xcolor}         
\usepackage{graphicx}
\usepackage{subcaption}
\usepackage{placeins}
\usepackage{multirow}
\usepackage{threeparttable}
\usepackage{tabularx}
\usepackage{enumitem}
\usepackage{algorithm}
\usepackage{algorithmic}

\newcommand{\crchange}[1]{#1}
\newcommand{\crdehedge}[1]{#1}

\IfFileExists{orcidlink.sty}{%
  \usepackage{orcidlink}
}{%
  \newcommand{\orcidlink}[1]{\href{https://orcid.org/#1}{[ORCID]}}
}

\AtBeginDocument{%
  \providecommand\BibTeX{{%
    \normalfont B\kern-0.5em{\scshape i\kern-0.25em b}\kern-0.8em\TeX}}}

\begin{document}

\title{PepSpecBench: A Unified Evaluation Benchmark for Peptide Tandem Mass Spectrometry Prediction}


\author{
  Zhiwen Yang\,$^{1,3,4}$\thanks{Equal contribution.} ,
  Pan Liu$^{1*}$,
  Yifan Li$^{1}$, Yunhua Zhong$^{1}$, 
  Jun Xia$^{1,2}$\thanks{Corresponding author.} \\
  $^{1}$The Hong Kong University of Science and Technology (Guangzhou).\\
  $^{2}$The Hong Kong University of Science and Technology.\\ 
  $^{3}$The Hong Kong Polytechnic University. \\
  $^{4}$Yangzhou University. \\
  \vspace{0.1cm} \\ 
  Corresponding author(s). E-mail(s): \href{mailto:junxia@hkust-gz.edu.cn}{junxia@hkust-gz.edu.cn}; \\
  Contributing authors: \href{mailto:zhiwen.yang@connect.polyu.hk}{zhiwen.yang@connect.polyu.hk}; \href{mailto:panliu@hkust-gz.edu.cn}{panliu@hkust-gz.edu.cn}; \\
  \href{mailto:yli994@connect.hkust-gz.edu.cn}{yli994@connect.hkust-gz.edu.cn}; \href{mailto:yunhuazhong@connect.hku.hk}{yunhuazhong@connect.hku.hk}
}

\maketitle


\begin{abstract}
Tandem mass spectrometry provides a high-throughput framework for identifying and quantifying proteins in complex biological samples. In computational proteomics, predicting peptide MS/MS spectra is a critical task, enabling downstream applications such as large-scale peptide identification and quantification. 
While deep learning architectures have substantially improved prediction accuracy, three evaluation challenges obscure the true progress of the field. First, inconsistent data preprocessing and incompatible model output spaces hinder fair model comparison. Second, flawed data splitting strategies can permit hidden sequence leakage and inflate reported performance. Third, existing evaluations typically lack systematic assessment of heterogeneous zero-shot domain shift and robustness to collision energy and precursor charge.
To address these challenges, we propose PepSpecBench, a unified benchmark for peptide MS/MS spectrum prediction. 
\crdehedge{The main leaderboard scores all six predictors on a shared b/y fragment-ion intensity subtask, so their outputs are directly comparable.} PepSpecBench standardizes data preprocessing across complementary public datasets, enforces a strict backbone-disjoint splitting strategy to prevent exact backbone overlap across partitions, and evaluates diverse architectures within a shared fragment-ion representation space. It further introduces a heterogeneous zero-shot domain-shift suite and metadata perturbation probes to assess robustness and whether models respond to fragmentation-relevant experimental conditions.
Across six representative models, we uncover changes in model rankings across training sources and external test subsets, along with differences in their responses to collision-energy and precursor-charge perturbations.
Code and data are available at \url{https://github.com/zwYang2004/PepSpecBench}.
\end{abstract}

\section{Introduction}
\label{sec:intro}


Tandem mass spectrometry (MS/MS) is the primary technology driving modern proteomics, enabling high-throughput identification and quantification of peptides in complex biological samples. MS/MS spectra are commonly interpreted through two complementary paradigms: database- or library-based matching, which is efficient and statistically well controlled but limited by the coverage of candidate sequence databases and empirical spectral libraries, and \textit{de novo} sequencing, which can operate without a fixed database but remains more challenging in accuracy and downstream integration~\cite{cox2008maxquant,yilmaz2024casanovo}. In this context, \textit{in silico} MS/MS spectrum prediction provides a practical bridge: by predicting fragment-intensity patterns for candidate peptides, researchers can construct and expand spectral libraries~\cite{fastspel2025,demichev2020diann}, improve data-independent acquisition (DIA) analysis~\cite{demichev2020diann}, and enhance peptide-spectrum match rescoring~\cite{kall2007percolator,yang2023msbooster}. In recent years, machine learning has provided state-of-the-art performance on this task, with models based on CNNs, RNNs, and Transformers mapping modified peptide sequences to their corresponding mass spectra~\cite{gessulat2019prosit,liu2020predfull,zeng2022alphapeptdeep,lapin2024unispec}.

\paragraph{\crchange{Background and benchmark scope.}}
\crchange{A mass spectrum records measured fragment intensity as a function of mass-to-charge ratio ($m/z$). A fragment ion is a charged peptide fragment produced by backbone cleavage; $b$ and $y$ ions are the complementary backbone-fragment series most consistently represented across the predictors evaluated here. A post-translational modification (PTM) is a residue or terminal chemical modification that changes peptide mass. We use peptide backbone to mean the amino-acid sequence after PTM annotations are stripped. The benchmark's primary task is fragment-ion intensity prediction for theoretical $b/y$ positions: models predict intensities at a predefined ion layout. This differs from full-spectrum prediction, which may also represent signal at non-canonical ions and across a dense $m/z$ grid. Forward spectrum prediction is operationally distinct from, and complementary to, reverse de novo sequencing.}

Three evaluation challenges hinder progress in peptide MS/MS prediction:

\begin{itemize}
    \item \textbf{The lack of standardized evaluation.} 
    Despite increasingly high reported prediction accuracies, results remain difficult to compare without a common evaluation protocol. Because mass spectrometry data is abundant and highly heterogeneous, researchers routinely apply customized filtering criteria, disparate post-translational modification (PTM) scopes, and ad hoc datasets for training and evaluation. Additionally, existing models predict fragment intensities in entirely incompatible native formats, ranging from 174-dimensional canonical ion vectors \cite{gessulat2019prosit} to 20,000-bin full-spectrum tensors \cite{liu2020predfull}. This inconsistency in datasets, output spaces and evaluation protocols precludes direct cross-model comparison and obscures true methodological advancements.

    \item \textbf{Leakage-prone data splitting.} 
   Leakage-prone partitioning severely compromises evaluation validity, an issue increasingly recognized in computational biology \cite{joeres2025datasail, bushuiev2024leakage, hermann2024leakage}. The key issue is the unit at which a split is performed: a row-level random split can place peptides sharing the exact same backbone but differing only by PTMs in both sets. Consequently, models can exploit memorized backbone patterns rather than learning fragmentation behavior for unseen backbones.

    
    \item \textbf{Limited robustness characterization.} 
    While recent efforts have introduced multi-species benchmarks for other proteomics tasks~\cite{wen2024multispecies,xu2025pep2prob}, fragment-intensity prediction is still typically evaluated within narrow in-domain settings. Existing evaluations also rarely probe whether models respond appropriately to fragmentation-relevant factors such as peptide length, normalized collision energy, and precursor charge, making it unclear whether high aggregate accuracy reflects physically grounded use of these conditioning variables.

\end{itemize}

\begin{figure*}[t]
    \centering
    \includegraphics[width=\textwidth]{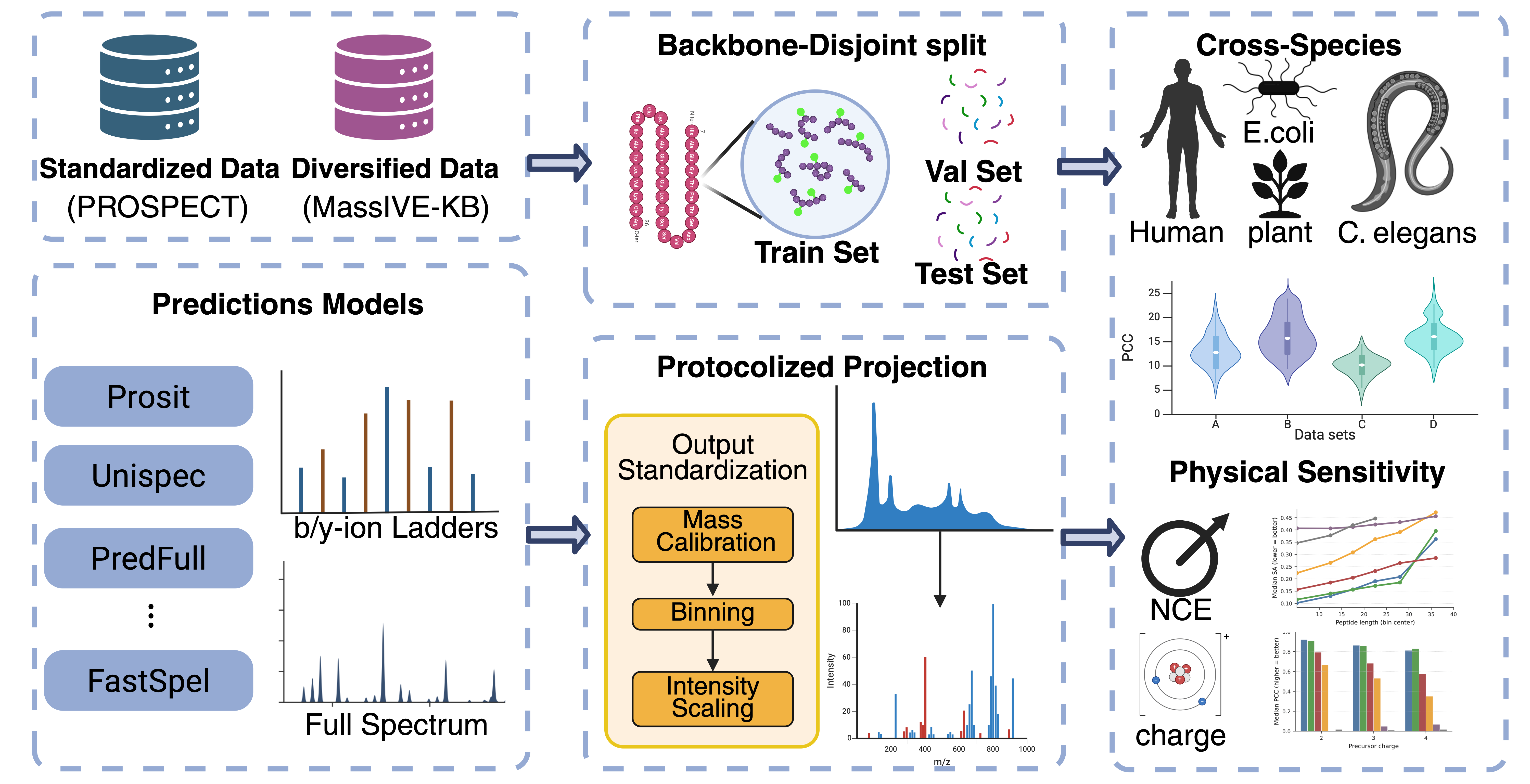}
   \caption{Overview of PepSpecBench. The framework consists of four stages: (1) standardized data integration, harmonizing PROSPECT and MassIVE-KB under a shared preprocessing scope; (2) leakage-aware data partitioning, enforcing a strict backbone-disjoint splitting strategy; (3) protocolized output projection, aligning diverse model predictions (ranging from discrete ion ladders to continuous full spectra) into a shared representation space; and (4) comprehensive robustness evaluation, assessing heterogeneous zero-shot domain shift and physical sensitivity.}
    \label{fig:overview}
\end{figure*}

To address these problems, we propose \textbf{PepSpecBench} (Figure~\ref{fig:overview}), a unified and leakage-aware benchmark for peptide MS/MS prediction. PepSpecBench brings datasets, model families, output spaces, and robustness tests into a single evaluation protocol.

Our work makes three main contributions. First, we harmonize PROSPECT~\cite{shouman2022prospect} and MassIVE-KB~\cite{wang2018assembling} under a shared data scope and align heterogeneous model outputs into a common fragment-ion representation, enabling direct comparison across six representative predictors spanning recurrent, convolutional, Transformer-based, dictionary-based, and lightweight architectures. Second, we introduce a strict backbone-disjoint splitting protocol to reduce sequence memorization across train and test sets; a controlled Prosit/PROSPECT ablation shows that row-level random splitting can yield a more optimistic estimate under this setting. Third, we extend evaluation beyond aggregate in-domain accuracy through heterogeneous zero-shot domain-shift testing, peptide-property stratification, and inference-time perturbation probes over collision energy and precursor charge, revealing that high in-domain accuracy does not necessarily imply robust transfer or sensitivity to conditioning variables.


\section{Related Work}
\label{sec:related}

\subsection{MS/MS Spectrum Prediction Models}
MS/MS spectrum prediction has evolved from feature-engineered and statistical approaches (MS2PIP~\cite{degroeve2013ms2pip} and its updated version~\cite{degroeve2019ms2pip}) to deep learning paradigms. By treating peptide fragmentation as a sequence-to-intensity mapping problem, diverse neural architectures have emerged to capture specific physical properties: RNNs capture sequential cleavage dependencies (pDeep~\cite{zeng2017pdeep}, Prosit~\cite{gessulat2019prosit}), CNNs extract local sequence motifs (PredFull~\cite{liu2020predfull}), and Transformers leverage self-attention to model long-range interactions (Prosit Transformer~\cite{ekvall2022prosit}, AlphaPeptDeep~\cite{zeng2022alphapeptdeep}). Additionally, efficient alternatives utilizing dictionary-based vocabulary learning (UniSpec~\cite{lapin2024unispec}) and ultra-fast linear approximations (FastSpel~\cite{fastspel2025}) have been developed. However, these models are developed independently with different PTM scopes and output spaces, predicting intensities in native formats. This structural heterogeneity precludes direct cross-model comparisons and obscures true advancements.

\subsection{Datasets and Benchmarking in Proteomics}
Modern in silico prediction relies heavily on large-scale public repositories (PRIDE~\cite{perez2025pride}, MassIVE-KB~\cite{wang2018assembling}) and curated spectral-library resources (ProteomeTools~\cite{zolg2017proteometools}, PROSPECT~\cite{shouman2022prospect}). While recent efforts have established multi-species benchmarks for parallel tasks such as \textit{de novo} sequencing (Casanovo~\cite{yilmaz2024casanovo}) and fragment-presence prediction~\cite{wen2024multispecies, xu2025pep2prob}, evaluation protocols for specific \textit{fragment-intensity prediction} remain severely outdated. Existing methods frequently utilize random or modified-sequence splitting strategies. As recently demonstrated across computational biology~\cite{joeres2025datasail, bushuiev2024leakage, hermann2024leakage}, such weak partitioning permits hidden homology leakage, allowing models to memorize structural backbones rather than learning generalizable fragmentation physics. Furthermore, current intensity-prediction evaluations are predominantly human-centric, leaving heterogeneous domain-shift behavior untested. A unified benchmark that jointly harmonizes output spaces, enforces strict backbone-disjoint leakage control, and evaluates out-of-distribution (OOD) generalization is urgently needed.


\section{Methods}
\label{sec:methods}

\subsection{Task and Benchmark Definition}
\label{subsec:task_def}

Let $\mathbf{x} = (s, z, e)$ denote an input triplet for peptide tandem mass spectrometry (MS/MS) prediction, where $s$ is a modified peptide sequence, $z \in \mathbb{Z}^{+}$ is the precursor charge state, and $e \in \mathbb{R}^{+}$ is the normalized collision energy (NCE).
A model $f_\theta$ maps $\mathbf{x}$ to a non-negative intensity vector that is projected to the shared canonical evaluation space $\mathcal{C}$ via a model-specific projection $\Pi_m$ defined in Section~\ref{subsec:baselines_alignment}: $\hat{\mathbf{I}}^{(c)} = \Pi_m(f_\theta(\mathbf{x}))$.
The benchmark target $\mathbf{I}^{(c)}$ is the experimentally observed fragment-intensity pattern projected to the same space $\mathcal{C}$. This task should be interpreted as a controlled evaluation of comparable backbone-fragment intensities. PepSpecBench restricts evaluation to peptides of length $L \in [6,\,40]$, precursor charge $z \in [1,\,6]$, and three common PTMs (acetylation, carbamidomethylation, and oxidation) which have reliable coverage across the selected public spectral-library sources (UNIMOD identifiers in ~\ref{app:data_pipeline}).
\crchange{Here, one sample denotes one retained spectrum/PSM row, including its peptide sequence, precursor charge, collision energy, and observed spectrum. \crdehedge{Reported medians are taken over retained test rows.}}



PepSpecBench combines two complementary in-domain sources.
\textbf{PROSPECT}~\cite{shouman2022prospect,gabriel2024prospect,zolg2017proteometools} is a curated spectral-library resource with broad PTM coverage and multiple NCE settings per peptide, making it well-suited for controlled physical-parameter analyses.

\textbf{MassIVE-KB}~\cite{wang2018assembling} is a large-scale, community-curated spectral library covering diverse human proteomics measurements. After filtering to the benchmark scope (peptides $L \in [6,40]$, charges $z \in [1,6]$, UNIMOD 1/4/35 PTMs), the dataset still encompasses highly heterogeneous sources. 
MassIVE-KB's specific harmonization is described in ~\ref{app:data_pipeline}.

Both sources are normalized to the shared physical scope and reconstructed into standardized 500k/50k/50k train/validation/test subsets.
Table~\ref{tab:benchmark_overview} summarizes the resulting benchmark components.

\begin{table}[t]
    \centering
    \caption{Descriptive statistics of the benchmark components used in the main paper. All components share the same peptide-length, charge, and PTM scope. For in-domain sources, train/validation/test percentages are reported separately in the same cell.}
    \label{tab:benchmark_overview}
    \setlength{\tabcolsep}{4pt}
    \resizebox{\columnwidth}{!}{%
    \begin{tabular}{@{}l l c c c@{}}
        \toprule
        \shortstack{\textbf{Component} \\ \strut} &
        \shortstack{\textbf{Domain} \\ \strut} &
        \shortstack{\textbf{Length} \\ \textbf{(Mean / Median)}} &
        \shortstack{\textbf{Charge} \\ \textbf{(Mean / Median)}} &
        \shortstack{\textbf{Spectra with PTMs} \\ \textbf{(\%)}} \\
        \midrule
        PROSPECT   & Human & 12.54 / 12.0 & 2.33 / 2.0 & 50.00 / 50.00 / 50.00 \\
        MassIVE-KB & Human & 18.70 / 17.0 & 2.81 / 3.0 & 24.60 / 27.17 / 27.91 \\
        \midrule
        \multirow{7}{*}{Heterogeneous OOD suite} & Human reference & 14.03 / 13.0 & 2.55 / 2.0 & 0.00 \\
         & Human HeLa & 21.01 / 19.0 & 2.33 / 2.0 & 36.01 \\
         & Yeast & 18.33 / 17.0 & 2.26 / 2.0 & 18.56 \\
         & E.\ coli reference & 15.50 / 14.0 & 2.36 / 2.0 & 0.00 \\
         & E.\ coli enriched & 19.53 / 17.0 & 2.27 / 2.0 & 25.52 \\
         & C.\ elegans & 15.25 / 14.0 & 2.89 / 3.0 & 0.00 \\
         & A.\ thaliana & 15.40 / 14.0 & 2.43 / 2.0 & 0.00 \\
        \bottomrule
    \end{tabular}}
\end{table}

\textbf{\crchange{Heterogeneous zero-shot domain-shift suite.}}
\crchange{To probe robustness beyond the human training domain, we curate a seven-subset heterogeneous zero-shot domain-shift suite spanning five species from ProteomeXchange studies~\cite{mueller2020proteome,ye2022carrier}.}
Each external subset is truncated to 20k spectra after applying the same filtering rules as the in-domain sources.
\crchange{All evaluation is conducted in a \emph{zero-shot} setting: models are trained exclusively on in-domain data and evaluated on external subsets without fine-tuning or adaptation. These subsets co-vary in species, PTM prevalence, peptide composition, NCE, acquisition source, instrument/fragmentation conditions, and sequence overlap.}
Detailed source provenance and distributional summaries are provided in ~\ref{app:data_pipeline}, \ref{app:ood_sources}, and~\ref{app:dataset_viz}.

\textbf{Backbone-Disjoint Split.}
A critical methodological requirement for MS/MS prediction benchmarks is the prevention of homology leakage: if a peptide backbone (the unmodified amino-acid sequence) appears in both the training and test sets, even with different PTM decorations, the model may exploit sequence memorization rather than learning transferable fragmentation physics.
We define the \emph{peptide backbone} $b(s)$ as the amino-acid sequence obtained by stripping all PTM annotations from $s$.
A partition $\{D_\mathrm{train}, D_\mathrm{val}, D_\mathrm{test}\}$ is \emph{Backbone-Disjoint} if the backbone sets of any two partitions are mutually exclusive.
This blocks both exact-sequence reuse and PTM-variant overlap across splits.

For PROSPECT, we enforce backbone disjointness by deterministic hashing of the peptide backbone into train/validation/test partitions.
For MassIVE-KB, we preserve the official disjoint splits, apply the same scope filtering within each partition, and empirically verify that the resulting train/validation/test sets maintain strict zero backbone overlap.

We contrast this protocol with two weaker alternatives: \emph{Naive Random Split} (each spectrum/PSM row is shuffled independently without regard to backbone identity) and \emph{Modified-Sequence Split} (splitting by the full modified peptide string, which still permits the same PTM-stripped backbone to appear with different PTMs across partitions), in a controlled ablation experiment (Section~\ref{subsec:split_ablation}).
Residual sequence similarity under Backbone-Disjoint splitting is quantified in ~\ref{app:backbone_similarity}.

\subsection{Baseline Models and Space Alignment}
\label{subsec:baselines_alignment}

We integrate six representative baselines that span the principal architectural families in MS/MS prediction. \emph{Sequence-centric models} predict b/y-ion intensities indexed by cleavage position and fragment charge: Prosit~\cite{gessulat2019prosit} (GRU encoder--decoder, $\sim$2.5M params), Prosit Transformer~\cite{ekvall2022prosit} (Transformer, $\sim$86M params), and AlphaPeptDeep~\cite{zeng2022alphapeptdeep} (modular Transformer, $\sim$1.8M params).
\emph{Full-spectrum and dictionary models} operate over broader ion spaces before projection: PredFull~\cite{liu2020predfull} (residual CNN with squeeze-excitation, $\sim$8.2M params) and UniSpec~\cite{lapin2024unispec} (attention-based dictionary model, $\sim$12M params).
\emph{Lightweight}: FastSpel~\cite{fastspel2025} ($\sim$50K parameters) is a linear model whose output is projected to the canonical layout, serving as a lower-bound reference.

All baselines are retrained on the benchmark data under the Backbone-Disjoint splits described above.
Model-specific adaptations, hyperparameters, and hardware settings are documented in ~\ref{app:model_adaptation} and~\ref{app:implementation}.

\textbf{Shared Canonical Evaluation Space.}
Because native output spaces are heterogeneous (sequence-indexed ion tensors, binned full spectra, fragment dictionaries, and dynamic-length vectors), raw model outputs are not directly comparable.
We address this by defining a \emph{protocolized projection} that maps every model's output to a single shared canonical evaluation space; all benchmark metrics are computed exclusively in this space. Let $\mathcal{C}$ denote the canonical ion space, which enumerates all combinations of cleavage position, ion type (b/y), and fragment charge up to a fixed maximum:
\begin{equation}
    \dim(\mathcal{C}) = (L_\mathrm{ref} - 1) \times 2 \times z_\mathrm{frag}^{\max},
    \label{eq:canonical_dim}
\end{equation}
where $L_\mathrm{ref}$ and $z_\mathrm{frag}^{\max}$ are the reference peptide-length and maximum fragment-charge parameters of the canonical space.
In the present benchmark, $L_\mathrm{ref} = 40$ and $z_\mathrm{frag}^{\max} = 3$, yielding $\dim(\mathcal{C}) = 234$.
In spectral-library prediction, $b/y$ ion ladders are the most consistently modeled peptide-backbone fragments across the evaluated predictors, making them a practical common target for heterogeneous model outputs.
For each model $m$, we define a projection function $\Pi_m$ that maps the model's native output to the canonical space:
\begin{equation}
    \hat{\mathbf{I}}^{(c)} = \Pi_m\!\bigl(f_\theta(\mathbf{x});\, \mathbf{x}\bigr) \in \mathcal{C}.
    \label{eq:projection}
\end{equation}
The projection $\Pi_m$ is model-specific at the input side, adapting to each model's native output format, but unified at the output side: all projections produce vectors in the same 234-dimensional space $\mathcal{C}$. Per-model projection details are provided in ~\ref{app:projection_details}. The resulting masked vectors are used for all metrics defined in Section~\ref{subsec:eval_protocol}; implementation details for model-specific and ground-truth projection are provided in ~\ref{app:projection_details}.
\crdehedge{Projection to the canonical space discards non-canonical signal: the canonical mask retains 19.8\% of the intensity that PredFull predicts on PROSPECT and 13.9\% under MassIVE-KB training (\ref{app:target_space_sensitivity}), because wide-output models such as PredFull and dictionary models such as UniSpec place signal on non-canonical ions that the mask discards. The same appendix reports how the ordering moves under expanded-ion and raw-20k targets.}

\subsection{Evaluation Protocols}
\label{subsec:eval_protocol}

Let $\hat{\mathbf{v}}, \mathbf{v} \in \mathbb{R}_{\ge 0}^{K}$ denote the predicted and observed intensity vectors after canonical projection and masking ($K = |\{j : \mathcal{M}_j = 1\}|$ valid positions).
\crchange{We report three complementary metrics: angular agreement (SA), its higher-is-better similarity form (SAS), and linear intensity correlation (PCC).}
\textbf{\emph{Spectral Angle} (SA)} quantifies the angular distance between two non-negative intensity vectors after $\ell_2$-normalization:
\begin{equation}
    \mathrm{SA}(\hat{\mathbf{v}}, \mathbf{v}) = \frac{1}{\pi}\arccos\!\left(\frac{\hat{\mathbf{v}}^\top \mathbf{v}}{\|\hat{\mathbf{v}}\|_2\;\|\mathbf{v}\|_2}\right),
    \label{eq:sa}
\end{equation}
where $\mathrm{SA} \in [0, 1]$ and lower is better. \textbf{\emph{Spectral Angle Similarity} (SAS)} $= 1 - \mathrm{SA}$.

\textbf{\emph{Pearson Correlation Coefficient} (PCC)} measures the linear agreement between predicted and observed intensities:
\begin{equation}
    \mathrm{PCC}(\hat{\mathbf{v}}, \mathbf{v}) = \frac{\sum_{j}(\hat{v}_j - \bar{\hat{v}})(v_j - \bar{v})}{\sqrt{\sum_{j}(\hat{v}_j - \bar{\hat{v}})^2 \sum_{j}(v_j - \bar{v})^2}},
    \label{eq:pcc}
\end{equation}
where $\bar{\hat{v}}$ and $\bar{v}$ are the masked-position means.
PCC captures whether relative intensity rankings are preserved, complementing the angular geometry measured by SA.
\crchange{SA is lower-is-better, whereas SAS $=1-\mathrm{SA}$ and PCC are higher-is-better.}

\textbf{Evaluation pipeline.}
For each test sample $\mathbf{x}$, we \textit{(i)} verify it satisfies the benchmark physical scope, \textit{(ii)} apply protocolized projection $\Pi_m$ to both prediction and ground truth, \textit{(iii)} construct the valid positional mask $\mathcal{M}(\mathbf{x})$, \textit{(iv)} compute SA, SAS, and PCC over masked positions, and \textit{(v)} aggregate via medians with bootstrap confidence intervals. \crchange{The physical-parameter probes interpret metadata use operationally: a model is considered condition-sensitive when its predictions respond to controlled changes in NCE or precursor charge \crdehedge{that affect fragmentation.}}


\label{sec:framework}  


\section{Experiments}
\label{sec:experiments}
\label{sec:results} 

\subsection{Unified Intra-Species Benchmarking}
\label{subsec:in_domain}


\begin{table*}[t]
    \centering
    \caption{In-domain benchmark in the shared canonical space (Backbone-Disjoint split, medians), with approximate model scale shown for context. Train (h) is wall-clock training hours; Inf. (ms) is median inference milliseconds per spectrum. Bootstrap CIs are reported in ~\ref{app:id_bootstrap_ci}.}
    \label{tab:id_main_stacked}
    \footnotesize
    \setlength{\tabcolsep}{1.2pt}
    \begin{tabular*}{\textwidth}{@{}l@{\hspace{4pt}}r@{\extracolsep{\fill}}cccccccccc@{}}
        \toprule
        & & \multicolumn{5}{c}{\textbf{MassIVE-KB}} & \multicolumn{5}{c}{\textbf{PROSPECT}} \\
        \cmidrule(lr){3-7} \cmidrule(lr){8-12}
        \textbf{Model} & \textbf{Params} & \textbf{SAS}$\uparrow$ & \textbf{SA}$\downarrow$ & \textbf{PCC}$\uparrow$
                        & \textbf{Train}$\downarrow$ & \textbf{Inf.}$\downarrow$
                        & \textbf{SAS}$\uparrow$ & \textbf{SA}$\downarrow$ & \textbf{PCC}$\uparrow$
                        & \textbf{Train}$\downarrow$ & \textbf{Inf.}$\downarrow$ \\
        \midrule
        Prosit Trans.     & 86M & \textbf{0.902} & \textbf{0.098} & \textbf{0.951} & 13.0 & 0.92 & 0.858 & 0.142 & 0.885 & 13.8 & 0.92 \\
        Prosit            & 2.5M & 0.901 & 0.100 & 0.949 & 9.2 & \textbf{0.06} & \textbf{0.862} & \textbf{0.138} & \textbf{0.892} & 4.7 & \textbf{0.05} \\
        AlphaPept.        & 1.8M & 0.871 & 0.129 & 0.915 & 4.4 & 0.20 & 0.722 & 0.278 & 0.593 & 9.1 & 0.55 \\
        PredFull          & 8.2M & 0.833 & 0.167 & 0.830 & 11.9 & 22.0 & 0.797 & 0.204 & 0.749 & 13.9 & 18.0 \\
        UniSpec           & 12M & 0.663 & 0.337 & 0.048 & 4.8 & 1.45 & 0.807 & 0.193 & 0.794 & 13.1 & 2.67 \\
        FastSpel          & 50K & 0.524 & 0.476 & 0.010 & \textbf{0.05} & 21.4 & 0.621 & 0.379 & 0.015 & \textbf{0.11} & 18.1 \\
        \bottomrule
    \end{tabular*}
\end{table*}

\textbf{Top rankings are source-dependent.}
Table~\ref{tab:id_main_stacked} summarizes in-domain performance in the shared canonical evaluation space under Backbone-Disjoint splitting. On MassIVE-KB, Prosit Transformer and Prosit form the top-performing group, with median SA values of 0.0977 and 0.0995, respectively, followed by AlphaPeptDeep (0.1288) and PredFull (0.1673). On PROSPECT, the ordering changes: Prosit achieves the lowest median SA (0.1376), while UniSpec (0.1933) ranks ahead of PredFull (0.2035) among the mid-tier models.
 
This dataset-dependent ordering is consistent with the composition shift between the two in-domain sources. MassIVE-KB contains longer peptides and higher precursor charges on average (~\ref{app:dataset_viz}), whereas PROSPECT contains shorter peptides with lower precursor charges on average. Charge-, PTM-, and length-stratified results for both datasets are reported in ~\ref{app:stratified}.

\textbf{Mid-tier models expose stronger cross-source instability.}
Below the top two, the mid-tier ordering also shifts across sources. On PROSPECT, UniSpec (SA 0.1933) slightly outperforms PredFull (0.2035), whereas on MassIVE-KB UniSpec deteriorates sharply to SA 0.3369 with near-zero PCC (0.048), while PredFull remains substantially stronger (SA 0.1673, PCC 0.830). This instability suggests a mismatch between UniSpec's learned dictionary vocabulary and the fragment-intensity patterns in MassIVE-KB. FastSpel remains the lower-bound reference on both datasets (SA 0.4757 on MassIVE-KB and 0.3792 on PROSPECT), confirming that peptide fragmentation cannot be captured adequately by a linear decomposition alone.
Fine-grained stratified analyses by charge, PTM type, and length are deferred to ~\ref{app:stratified}.

\subsection{\crchange{Controlled Split Sensitivity}}
\label{subsec:split_ablation}
\crchange{In a controlled Prosit/PROSPECT ablation, row-level random splitting gives a more optimistic estimate than the backbone-disjoint protocol (median SA 0.1254 vs. 0.1376; PCC 0.9099 vs. 0.8921). The direction is reproduced at a second seed (SA 0.1282 vs. 0.1391), while the modified-sequence split remains close to backbone-disjoint (SA 0.1392). The associated bootstrap intervals, overlap audit, and stricter-split stress tests are reported in \ref{app:split_sensitivity}. \crdehedge{These splits jointly change overlap at the spectrum-row, modified-sequence, and backbone levels.}}

%

\subsection{\crchange{Heterogeneous Zero-Shot Domain Shift}}
\label{subsec:cross_species}

\begin{table*}[t]
    \centering
    \caption{\crchange{Heterogeneous zero-shot domain-shift results. Entries show \texttt{SA\,/\,SAS} medians; E.\ coli is aggregated over both sources. The complete seven-subset audit is in \ref{tab:ood_full}.}}
    \label{tab:ood_main}
    \footnotesize
    \setlength{\tabcolsep}{2pt}
    \begin{tabular*}{\textwidth}{@{\extracolsep{\fill}}lcccccc@{}}
        \toprule
        \multicolumn{7}{c}{\textbf{MassIVE-KB Trained}} \\
        \midrule
        \textbf{Model} & \textbf{ID (SA)} & \textbf{E.\ coli} & \textbf{C.\ elegans} & \textbf{A.\ thaliana} & \textbf{Yeast} & \textbf{HeLa} \\
        \midrule
        Prosit            & 0.100 & 0.476/0.524 & 0.471/0.529 & 0.484/0.516 & 0.477/0.523 & 0.471/0.529 \\
        Prosit Tran.      & \textbf{0.098} & 0.480/0.520 & 0.472/0.528 & 0.485/0.515 & 0.481/0.519 & 0.475/0.525 \\
        PredFull          & 0.167 & \textbf{0.238}/\textbf{0.762} & \textbf{0.222}/\textbf{0.778} & \textbf{0.203}/\textbf{0.797} & \textbf{0.281}/\textbf{0.719} & \textbf{0.220}/\textbf{0.780} \\
        AlphaPep.         & 0.129 & 0.473/0.527 & 0.470/0.530 & 0.482/0.518 & 0.473/0.527 & 0.468/0.532 \\
        UniSpec           & 0.337 & 0.388/0.612 & 0.382/0.618 & 0.384/0.616 & 0.393/0.607 & 0.380/0.620 \\
        FastSpel          & 0.476 & 0.465/0.535 & 0.476/0.524 & 0.469/0.531 & 0.461/0.539 & 0.461/0.539 \\
        \midrule
        \multicolumn{7}{c}{\textbf{PROSPECT Trained}} \\
        \midrule
        \textbf{Model} & \textbf{ID (SA)} & \textbf{E.\ coli} & \textbf{C.\ elegans} & \textbf{A.\ thaliana} & \textbf{Yeast} & \textbf{HeLa} \\
        \midrule
        Prosit            & \textbf{0.138} & 0.286/0.714 & \textbf{0.160}/\textbf{0.840} & \textbf{0.094}/\textbf{0.906} & 0.284/0.716 & 0.244/0.756 \\
        Prosit Tran.      & 0.142 & 0.298/0.702 & 0.169/0.831 & 0.102/0.898 & 0.288/0.712 & 0.259/0.741 \\
        PredFull          & 0.203 & \textbf{0.236}/\textbf{0.764} & 0.199/0.801 & 0.163/0.837 & \textbf{0.236}/\textbf{0.764} & \textbf{0.196}/\textbf{0.804} \\
        AlphaPep.         & 0.278 & 0.312/0.688 & 0.246/0.754 & 0.157/0.843 & 0.319/0.681 & 0.274/0.726 \\
        UniSpec           & 0.193 & 0.299/0.701 & 0.188/0.812 & 0.120/0.880 & 0.294/0.706 & 0.251/0.749 \\
        FastSpel          & 0.379 & 0.369/0.631 & 0.392/0.609 & 0.344/0.656 & 0.370/0.629 & 0.361/0.639 \\
        \bottomrule
    \end{tabular*}
\end{table*}

\textbf{\crchange{OOD transfer changes the in-domain ordering under heterogeneous domain shift.}}
\crchange{We evaluate seven external test subsets in a zero-shot setting, including four non-human species, a human reference subset, and HeLa as an intra-species shift. The subsets also differ in PTM load, NCE, acquisition source, \crdehedge{instrument/fragmentation, and sequence overlap.} Table~\ref{tab:ood_main} reports the main results.}

\textbf{\crchange{Training-source and architecture jointly shape OOD behavior.}}
\crchange{Under MassIVE-KB training, PredFull has lower SA than the sequence-centric models on the external subsets, while under PROSPECT training the ordering varies by subset. \crdehedge{Negative PCC on some subsets reflects anti-correlation in the canonical coordinates, which occurs when predicted intensity mass is distributed differently from the library spectrum.}}

\textbf{\crchange{\crdehedge{The OOD advantage on} A.\ thaliana is covariate-sensitive.}}
\crchange{The lower aggregate SA for Prosit on \textit{A.\ thaliana} (0.094 versus PROSPECT ID 0.138) is consistent with a covariate-shift decomposition: after matching on NCE and PTM, the PROSPECT reference subset has SA 0.070 versus 0.094 OOD (\ref{app:ood_covariate_shift}).}

\textbf{OOD robustness is architecture--data dependent.}
\crchange{PredFull’s external behavior is not uniformly superior: under the intra-species shift to HeLa, it shows a more pronounced degradation in angular metrics, while sequence-centric models exhibit larger drops in PCC. \crdehedge{One reading is that its full-spectrum formulation preserves local mass--intensity correlations that survive this shift; separating that from the PTM and instrument differences between the subsets would require matched acquisition, which public spectral libraries do not provide.}}
In-domain bootstrap confidence intervals are in ~\ref{app:id_bootstrap_ci}.

\subsection{Analysis Across Peptide Properties}
\label{subsec:peptide_properties}

\begin{figure*}[t]
    \centering
    \includegraphics[width=1\textwidth]{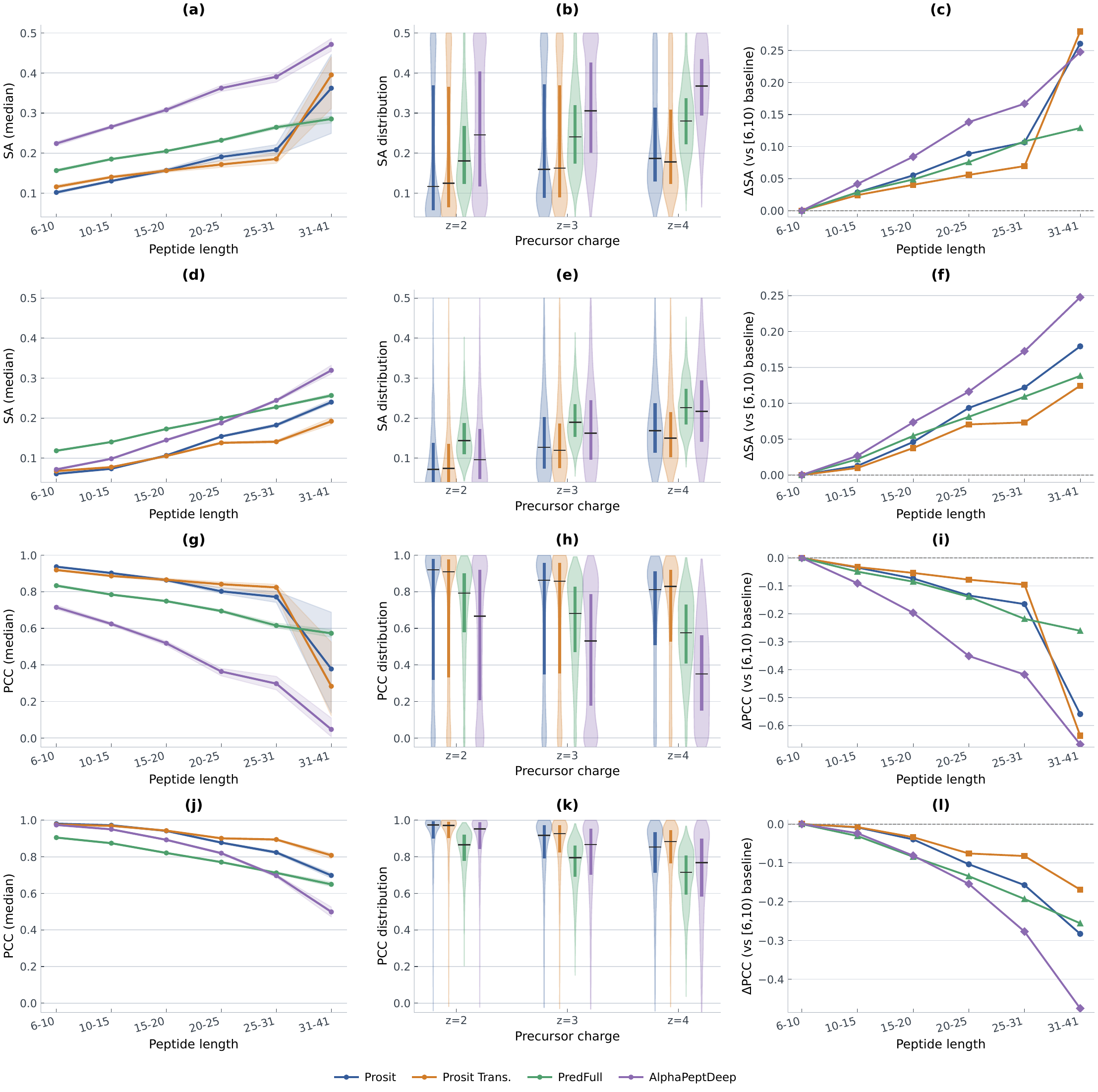}
    \caption{Peptide-property analysis in the shared canonical space. Rows: PROSPECT SA, MassIVE-KB SA, PROSPECT PCC, MassIVE-KB PCC. Columns: peptide length, charge, $\Delta$SA/$\Delta$PCC by length (baseline = [6,10) bin). Bands: bootstrap CIs.}
    \label{fig:delta_sa_decay}
    \vspace{-6pt}
\end{figure*}

To isolate the impact of peptide properties, we stratify prediction errors for Prosit, Prosit Transformer, PredFull, and AlphaPeptDeep by length, precursor charge, and NCE. Figure~\ref{fig:delta_sa_decay} shows the length and charge results; NCE results are in ~\ref{app:stratified}.

\textbf{Longer peptides are harder to predict.}
As length increases, peptides generate more cleavage sites, more fragment-charge combinations, and higher-order intensity interactions. On PROSPECT, Prosit's median SA rises from about 0.10 on short peptides to about 0.21 for $L>25$, while median PCC falls from 0.9367 to 0.7715; the corresponding MassIVE-KB trend spans roughly 0.06 to 0.18. 
Figure~\ref{fig:delta_sa_decay} quantifies this as the change in SA from the [6,10) reference length bin.

\textbf{Higher precursor charge increases prediction difficulty.}
On MassIVE-KB, Prosit Transformer's error increases from SA 0.075 at $z{=}2$ to 0.150 at $z{=}4$, whereas Prosit rises from 0.072 to 0.168; on PROSPECT, Prosit similarly increases from 0.116 to 0.186. These charge-stratified gaps mechanistically explain Prosit Transformer's slight advantage on MassIVE-KB, which contains more highly charged peptides.

\textbf{Higher NCE amplifies fragmentation complexity.}
Higher NCE bins on PROSPECT produce worse predictions under both SA and PCC, consistent with the physical expectation that more energetic fragmentation generates more complex and noisy spectra. For Prosit, median SA rises from 0.0806 to 0.1974 and median PCC drops from 0.9616 to 0.7888 between the 0.20--0.25 and 0.30--0.35 bins. MassIVE-KB occupies a single NCE bin after reconstruction (~\ref{app:stratified}).


\subsection{Physical Parameter Sensitivity}
\label{subsec:model_sensitivity}

We probe physical grounding through inference-time NCE and precursor-charge perturbations.

\begin{figure}[t]
    \centering
    \begin{subfigure}[t]{0.3\columnwidth}
        \centering
        \includegraphics[width=\linewidth]{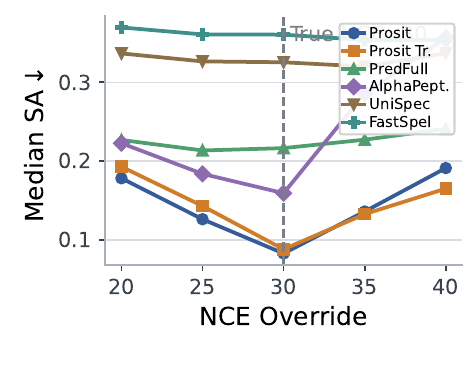}
        \caption{}
        \label{fig:nce_sensitivity_a}
    \end{subfigure}
    \hfill
    \begin{subfigure}[t]{0.33\columnwidth}
        \centering
        \includegraphics[width=\linewidth]{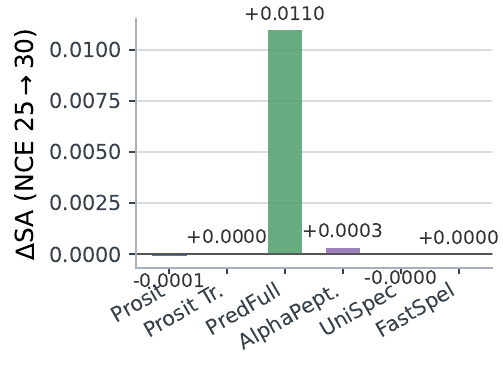}
        \caption{}
        \label{fig:nce_sensitivity_b}
    \end{subfigure}
    \hfill
    \begin{subfigure}[t]{0.33\columnwidth}
        \centering
        \includegraphics[width=\linewidth]{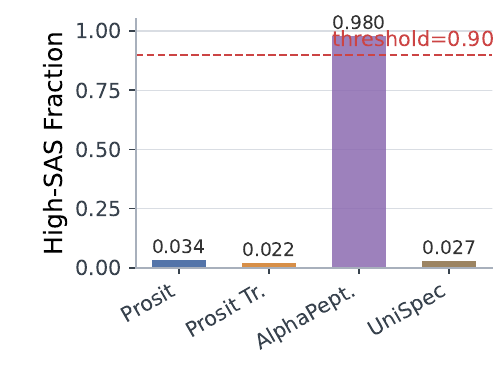}
        \caption{}
        \label{fig:nce_sensitivity_c}
    \end{subfigure}
    \caption{Physical-parameter sensitivity: three experiments side-by-side. (a) NCE Calibration Analysis on PROSPECT: median SA vs.\ overridden NCE; dashed line = true NCE${}=30$. (b) Blind NCE Perturbation on MassIVE-KB: $\Delta$SA from NCE 25 to 30. (c) Charge-State Perturbation on PROSPECT-Mini: fraction of spectra with SAS$>0.90$ between predictions at $z{=}2$ and $z{=}3$.}
    \label{fig:nce_sensitivity_fullrun}
\end{figure}

\textbf{NCE calibration depends on multi-NCE training.}
We use two NCE probes: an inference-time NCE sweep on PROSPECT and a blind NCE shift (from 25 to 30) on MassIVE-KB (Figure~\ref{fig:nce_sensitivity_fullrun} (a) -- (b), ~\ref{app:nce_sensitivity}). Under the \emph{NCE Calibration Analysis} on PROSPECT, the Prosit family correctly achieves minimal error at the true NCE (30), whereas PredFull and UniSpec misalign their minima. Conversely, the \emph{Blind NCE Perturbation} on MassIVE-KB reveals that models trained on effectively single-NCE data largely ignore NCE metadata, exhibiting near-zero prediction shifts. This confirms that robust NCE calibration requires explicit multi-NCE training distributions.

\textbf{Some models ignore charge as a conditioning variable.}
We force an artificial charge shift ($z{=}2{\rightarrow}3$) on identical peptides at inference time and measure the fraction of collapsed predictions, defined by SAS$>0.90$ between the original and charge-shifted predictions (Figure~\ref{fig:nce_sensitivity_fullrun} (c) ). This perturbation exposes charge-ignoring failures. While Prosit, UniSpec and Prosit Transformer correctly adjust their predictions (retaining $<$4\% collapse), AlphaPeptDeep suffers a massive collapse rate (98.0\%), over-relying on sequence features while discarding the physically critical charge scalar (~\ref{app:alphapeptdeep}). Table~\ref{tab:charge_mode_collapse_fullrun} reports the full distributions.

\FloatBarrier

\section{Conclusion and Limitations}
\label{sec:conclusion}

\crchange{PepSpecBench compares six predictors on a shared $b/y$ target. Row-level random splits yield optimistic Prosit/PROSPECT estimates; heterogeneous zero-shot shifts change rankings; and high accuracy does not guarantee sensitivity to collision energy or precursor charge.}

\textbf{\crchange{Limitations and future work.}} \crchange{\crdehedge{Rankings differ under expanded-ion and raw-20k targets (~\ref{app:target_space_sensitivity}).} Future versions should expand to rarer modifications, broader instrument families, and additional acquisition settings. Backbone-disjoint splitting removes exact overlap but leaves higher-order sequence homology, motivating homology-aware partitioning. Confounding between NCE and instrument type remains difficult to resolve using public libraries alone.}

\textbf{Broader impact.} PepSpecBench supports spectral-library expansion, DIA analysis, and peptide-spectrum match rescoring by identifying leakage and testing generalization across biological and experimental settings. Misuse of its rankings could lead to inappropriate model selection if users overlook instrument configuration, training-source bias, downstream application requirements, or differences between canonical- and native-output rankings.

\clearpage
\begin{ack}
We sincerely thank the anonymous reviewers for their insightful comments and constructive suggestions. This research is supported by the National Natural Science Foundation of China Project (No.~62606430, No.~623B2086), sponsored by CCF-GHFund (No.~OF 2026005), and sponsored by CIPS-SMP-Zhipu Large Model Fund, Ant Group and Shanghai AI Lab. The authors declare no competing interests.
\end{ack}

\bibliographystyle{unsrtnat}
\bibliography{PepSpecBench}

\appendix
\newpage
\renewcommand{\thesection}{Appendix~\Alph{section}}

\section{Datasheet for Datasets}
\label{app:datasheet}
Following the NeurIPS Datasets and Benchmarks track recommendations~\cite{gebru2021datasheets}, we provide a Datasheet for PepSpecBench.

\paragraph{Motivation.} To enable rigorous, reproducible evaluation of MS/MS prediction models with consistent splits, PTM handling, and leakage control.

\paragraph{Composition.} (1) PROSPECT-Mini: 500k/50k/50k spectra (train/val/test) from ProteomeTools, balanced Unmod/PTM. (2) MassIVE-KB-Mini: 500k/50k/50k from MassIVE-KB v1 human subset, natural PTM distribution. (3) PepSpecBench-OOD: 7 subsets $\times$ 20k spectra covering five species (human, yeast, \textit{E. coli}, \textit{C. elegans}, \textit{A. thaliana}).

\paragraph{Collection.} PROSPECT and MassIVE-KB are public spectral libraries; we apply filtering ($L \in [6,40]$, $z \in [1,6]$, UNIMOD 1/4/35) and backbone-level splitting. OOD data from ProteomeXchange with MaxQuant~\cite{cox2008maxquant} 0.01 FDR.

\paragraph{Preprocessing.} PTM normalization (regex $\to$ canonical UNIMOD), scope filters, deterministic sampling. See Section~\ref{app:data_pipeline}.

\paragraph{Distribution.} PROSPECT-Mini is exactly 50.00\% Unmod / 50.00\% PTM in all splits by construction. MassIVE-KB-Mini preserves the natural distribution (Train/Val/Test Unmod = 75.40\%/72.83\%/72.09\%). \crchange{OOD subsets are source-specific 20k-spectrum mini tables with heterogeneous PTM rates ranging from 0\% in the core proteome subsets to 36.01\% in the HeLa trypsin enriched subset.}

\paragraph{Biases.} Human-centric training data; instrumentation bias (Q Exactive family dominant); backbone split does not eliminate homology-level leakage (see Limitations).

\paragraph{Use.} Benchmarking MS/MS prediction models; no identification of individuals. Intended for research and method development.

\section{Dataset Visualization}
\label{app:dataset_viz}
\begin{figure}[h]
    \centering
    \includegraphics[width=0.95\columnwidth]{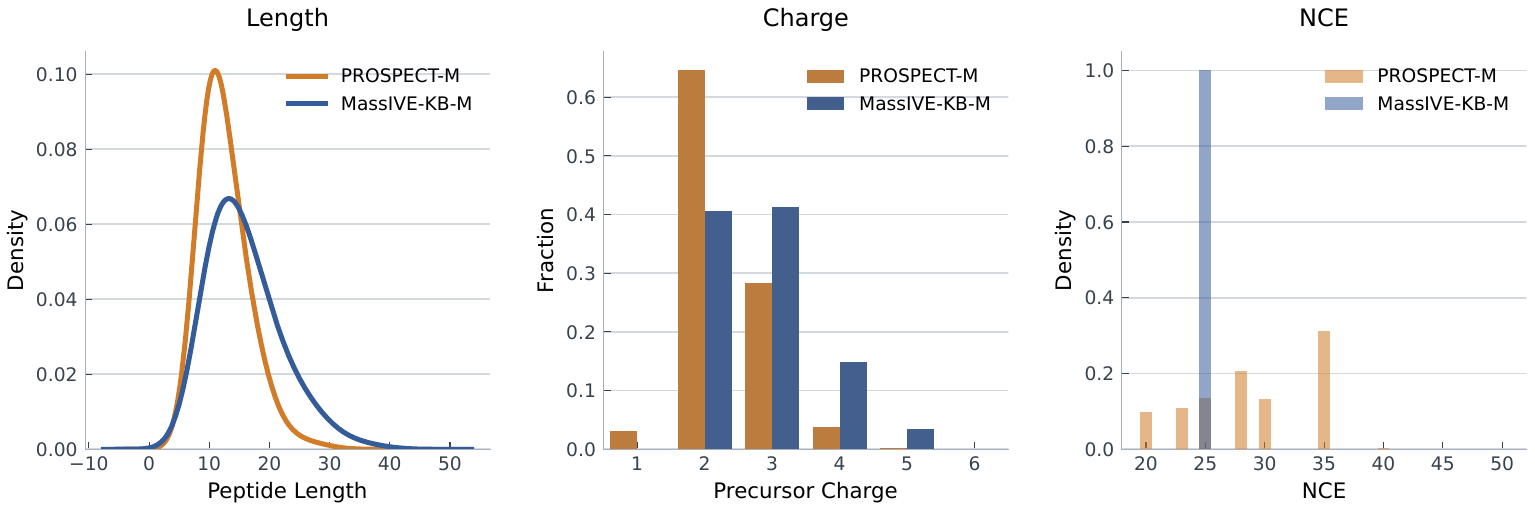}
    \caption{Measured mini-corpus distribution comparison aggregated over train/validation/test. Left: peptide-length KDE. Middle: precursor-charge distribution. Right: NCE histogram. PROSPECT-Mini spans multiple NCE settings, whereas MassIVE-KB-Mini is concentrated at NCE=25.}
    \label{fig:dataset_length_charge}
\end{figure}

\begin{figure}[t]
    \centering
    \includegraphics[width=0.7\columnwidth]{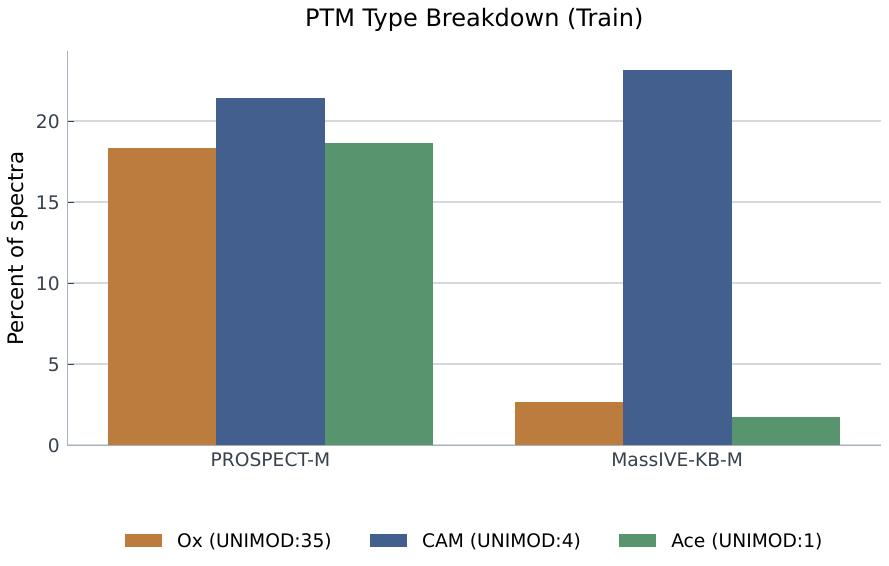}
    \caption{PTM-type breakdown on training splits (Ox/CAM/Ace). PROSPECT has balanced representation; MassIVE-KB is dominated by Carbamidomethylation (Cys).}
    \label{fig:dataset_ptm_breakdown}
\end{figure}

Figures~\ref{fig:dataset_length_charge}--\ref{fig:dataset_ptm_breakdown} provide visual summaries of the final PepSpecBench-Mini datasets, complementing the benchmark summary in Table~\ref{tab:benchmark_overview}.

\subsection{Homology Leakage: Train--Test Backbone Similarity}
\label{app:backbone_similarity}
To quantify residual homology-level leakage under our backbone-disjoint split, we compute the minimum Levenshtein edit distance from each test backbone to the nearest train backbone. For the PROSPECT dataset, we empirically find a mean minimum edit distance of 1.87 and a median of 2.0. The fraction of test backbones with edit distance 0 (exact match in train) is 0.0\%, confirming the efficacy of our backbone-disjoint protocol. However, 13.0\% of test backbones have an edit distance of $\le 1$ (single-amino-acid variants), indicating that residual sequence similarity persists. \crchange{This motivates the heterogeneous zero-shot domain-shift suite, \crdehedge{which evaluates robustness under larger sequence and acquisition shifts.}}
MassIVE-KB backbone similarity is not computed because its official split already ensures backbone disjointness, which we verified empirically.

\section{Data Processing and Reconstruction Pipeline}
\label{app:data_pipeline}

\subsection{Raw Data Acquisition}

We start from publicly available spectral libraries and their associated identifications:
\begin{itemize}[leftmargin=*]
    \item \textbf{PROSPECT.} We download the PROSPECT benchmark data~\cite{shouman2022prospect,gabriel2024prospect} and use the authors' released library-level identifications as our starting point, rather than re-running peptide identification.
    \item \textbf{MassIVE-KB.} We obtain the MassIVE-KB v1 spectral library~\cite{wang2018assembling} and its human-only subset, again reusing the official identifications provided by the library authors.
\end{itemize}

All downstream processing in PepSpecBench operates on table-like representations (parquet files) derived from these libraries, not on raw vendor files.

\subsection{Canonicalization and Global Filtering}

To harmonize the dual-source corpus, we apply a shared set of quality and scope constraints before any splitting or sampling. This stage produces PTM-normalized tables for both source corpora.

\paragraph{Quality control and basic filters.}
For both corpora we:
\begin{itemize}[leftmargin=*]
    \item Restrict precursor charge to $z \in [1, 6]$ and peptide length to $L \in [6, 40]$ amino acids.
    \item Retain only high-confidence spectra (e.g., Andromeda score $\ge 70$ and mass error within 20~ppm for PROSPECT), following the original benchmark recommendations.
    \item Drop spectra with missing or malformed metadata required by downstream models (charge, collision energy, raw file identifier, etc.).
\end{itemize}

\paragraph{PTM normalization.}
We standardize PTM annotations to a canonical UNIMOD vocabulary:
\begin{itemize}[leftmargin=*]
    \item Parse modified peptide strings with a regex-based tokenizer and map any recognized PTM to one of the three benchmark PTMs: Acetylation (UNIMOD:1), Carbamidomethylation (UNIMOD:4), and Oxidation (UNIMOD:35).
    \item Discard peptides containing unsupported or ambiguous PTMs to maintain a clean, comparable label space.
    \item For each remaining peptide, derive (i) a \emph{normalized sequence} where all PTM tags are rewritten as canonical \texttt{[UNIMOD:x]} tokens; (ii) a \emph{naked sequence} obtained by stripping all PTM tags; and (iii) auxiliary columns such as \texttt{has\_ptm} and a coarse \texttt{ptm\_bucket} (Ox/CAM/Ace vs. unmodified).
\end{itemize}

This produces a pair of harmonized, PTM-cleaned tables that serve as the sole inputs to the reconstruction and mini-sampling stage, without imposing any particular on-disk layout.

\subsection{Backbone-Level Splitting}

We enforce strict disjointness at the level of the naked peptide sequence.

\paragraph{PROSPECT.}
For PROSPECT, we ignore any existing split markers and re-partition the data using a deterministic hash on the naked sequence:
\begin{itemize}[leftmargin=*]
    \item Compute $h = \text{MD5}(\text{naked sequence})$ and map it to a bucket in $[0,99]$.
    \item Assign buckets $0$--$79$ to Train, $80$--$89$ to Validation, and $90$--$99$ to Test, yielding an 80/10/10 split by backbone (8:1:1).
\end{itemize}
This guarantees that no naked sequence can appear in more than one split, even if it occurs with different PTM patterns.

\paragraph{MassIVE-KB.}
For MassIVE-KB, we preserve the official Train/Validation/Test directories released by the library authors and \emph{never} move peptides across these boundaries. All subsequent downsampling is carried out independently within each official split, ensuring that we do not weaken the original disjointness guarantees.

\subsection{Construction of PepSpecBench-Mini}

Given the backbone-level splits above, we construct the standardized mini subsets using a deterministic streaming sampler over the filtered tables.

\paragraph{PROSPECT (balanced).}
Within each PROSPECT split we:
\begin{itemize}[leftmargin=*]
    \item Apply the global filters and PTM normalization described above to each streamed batch.
    \item For every candidate spectrum, compute a deterministic \emph{sampling key} by hashing the tuple (naked sequence, precursor charge, collision energy, seed). This key defines a total order that is stable across machines.
    \item Accumulate candidates in four buckets per split: unmodified, Ox (UNIMOD:35), CAM (UNIMOD:4), and Ace (UNIMOD:1).
    \item From these buckets, select a fixed number of spectra per split (500k/50k/50k for Train/Validation/Test in our experiments), enforcing approximately 50\% unmodified and 50\% modified peptides. Among modified peptides, we further draw approximately equal proportions from the Ox/CAM/Ace buckets without oversampling. Within each bucket, spectra are sorted by the sampling key and the first $N$ are taken (deterministic, reproducible).
\end{itemize}
The resulting tables form the PROSPECT component of PepSpecBench-Mini.

\paragraph{MassIVE-KB (natural).}
For MassIVE-KB we instead preserve the empirical PTM and charge distributions:
\begin{itemize}[leftmargin=*]
    \item Process each official split directory independently, streaming over its constituent parquet files in a fixed, deterministic order.
    \item Apply the same global filters and PTM normalization as for PROSPECT, but \emph{do not} re-balance PTM categories.
    \item Within each split, perform a simple top-$N$ truncation based on the streaming order (files processed in a fixed, deterministic order) to obtain 500k/50k/50k spectra for Train/Validation/Test, respectively.
\end{itemize}
In MassIVE-KB, we assign a unified default normalized collision energy (NCE) of 25 to all retained MassIVE-KB spectra during reconstruction. This harmonization step ensures consistent downstream model inputs and makes MassIVE-KB effectively single-bin in NCE.
These mini-splits constitute the MassIVE-KB component of PepSpecBench-Mini. For PROSPECT, disjointness is guaranteed by the hash-based partition; for MassIVE-KB, we preserve the official splits. The resulting split sizes are summarized in Table~\ref{tab:benchmark_overview}.

\subsection{Quick Random/Sequence Split Construction for Ablation}
\label{app:quick_split_construction}
For the split-ablation experiment in Section~\ref{subsec:split_ablation}, we additionally build two deterministic quick split variants from reconstructed PROSPECT data:
\begin{itemize}[leftmargin=*]
    \item \textbf{Input pool.} Concatenate all rows from the reconstructed PROSPECT train/validation/test tables.
    \item \textbf{Random split.} Compute an MD5 bucket from a row-level key (\texttt{sample\_key} plus row index when available), then assign 80/10/10 to Train/Val/Test.
    \item \textbf{Modified-sequence split.} Compute the bucket from \texttt{modified\_sequence}, then apply the same 80/10/10 rule.
    \item \textbf{Implementation.} Generated by the benchmark split-construction utility used in this work.
\end{itemize}
The realized quick-split datasets are stored as reconstructed benchmark tables with deterministic compatibility aliases and approximately 600k spectra each (80/10/10 ratio).
This quick protocol is used only for split-ablation sensitivity analysis and does not replace the main benchmark split definitions.

\subsection{\crchange{Construction of the Heterogeneous Zero-Shot Domain-Shift Suite}}
\label{app:ood_sources}

\crchange{To evaluate heterogeneous zero-shot domain shift under the same PTM scope as the in-domain benchmarks, we construct a seven-subset OOD suite spanning five species and multiple acquisition contexts. Each source is processed independently to preserve domain identity while producing compact, reproducible evaluation subsets.}

\paragraph{Data sources.}
Four datasets (human, \textit{E. coli}, \textit{C. elegans}, \textit{A. thaliana}) are derived from PXD014877~\cite{mueller2020proteome}, a large-scale multi-species proteomics study covering 100 organisms. Protein sequence databases were obtained from UniProt KB reference proteomes: \textit{H. sapiens} (UP000005640), \textit{A. thaliana} (UP000006548), \textit{C. elegans} (UP000001940), and \textit{E. coli} (UP000000625). Searches were performed with MaxQuant~\cite{cox2008maxquant} under 0.01 FDR control. Three additional datasets (\textit{S. cerevisiae} yeast, \textit{E. coli}, and human HeLa trypsin) are drawn from PXD027742~\cite{ye2022carrier}, a study on carrier proteome effects in single-cell proteomics using TMTpro-labeled mixed-species samples. Table~\ref{tab:ood_metadata_summary} summarizes the origin and experimental setup of each OOD subset.
For reproducibility, the Sample Size and PTM\% columns in Table~\ref{tab:ood_metadata_summary} are computed directly from the finalized OOD mini tables after the same filtering and deterministic top-20k truncation used in evaluation.

\begin{table}[h]
    \centering
\caption{\crchange{OOD metadata audit: PepSpecBench heterogeneous-domain-shift mini suite. PXD = ProteomeXchange. PTM\% = fraction of spectra containing UNIMOD 1/4/35 after filtering and deterministic top-20k truncation.}}
    \label{tab:ood_metadata_summary}
    \resizebox{\columnwidth}{!}{%
    \begin{tabular}{lllllcc}
        \toprule
        \textbf{Source} & \textbf{Species} & \textbf{Instrument} & \textbf{Fragmentation} & \textbf{Sample Size} & \textbf{PTM\%} \\
        \midrule
        homo\_sapiens\_with\_seq & \textit{H. sapiens} & Q Exactive HF & HCD & 20{,}000 & 0.00 \\
        coli\_with\_seq & \textit{E. coli} & Q Exactive HF & HCD & 20{,}000 & 0.00 \\
        c\_elegans\_with\_seq & \textit{C. elegans} & Q Exactive HF & HCD & 20{,}000 & 0.00 \\
        a\_thaliana\_with\_seq & \textit{A. thaliana} & Q Exactive HF & HCD & 20{,}000 & 0.00 \\
        \midrule
        Yeast & \textit{S. cerevisiae} & Q Exactive & HCD & 20{,}000 & 18.56 \\
        Ecoli & \textit{E. coli} & Q Exactive & HCD & 20{,}000 & 25.52 \\
        HeLa\_trypsin & \textit{H. sapiens} (HeLa) & Q Exactive & HCD & 20{,}000 & 36.01 \\
        \bottomrule
    \end{tabular}%
    }
\end{table}
\noindent\textbf{ProteomeXchange provenance.} Four datasets (human, \textit{E. coli}, \textit{C. elegans}, \textit{A. thaliana}) from PXD014877~\cite{mueller2020proteome}; three (Yeast, E. coli, HeLa) from PXD027742~\cite{ye2022carrier}. After the final mini reconstruction, the four PXD014877 subsets remain PTM-free under our UNIMOD 1/4/35 scope, whereas the three PXD027742 subsets preserve the intended PTM-enriched contexts.

\paragraph{Global filters.}
For every OOD source table, we enforce the same scope constraints as PepSpecBench: peptide length $L \in [6, 40]$ and precursor charge $z \in [1, 6]$.

\paragraph{PTM normalization and whitelisting.}
We normalize modified sequences using the same regex-based tokenizer as our main reconstruction procedure. Any bracket token is mapped to one of the benchmark PTMs (UNIMOD:1/4/35). Spectra containing any other modification are discarded to ensure model compatibility and a consistent evaluation scope.

\paragraph{Deterministic ranking and truncation.}
To obtain a compact yet reproducible OOD evaluation set, we compute a deterministic ranking key for each spectrum:
$\text{key} = \mathrm{MD5}(\text{naked sequence} \| \text{charge} \| \text{seed})$, with a fixed seed of 42. For each source file we then select the top-20{,}000 spectra by this key, yielding seven fixed-size evaluation subsets.

\paragraph{Projection to shared canonical space.}
For all models, we convert each OOD parquet table to the shared canonical evaluation space $\mathcal{C}$ using the same ion-labeling utility as the main benchmark, producing row-aligned outputs.

\paragraph{Scope alignment and PROSPECT OOD.}
All three phases---training, in-domain test, and OOD evaluation---use the same physical scope. Models trained on PROSPECT are evaluated on the same OOD suite summarized in Table~\ref{tab:ood_main}.

\subsection{Complete OOD Results (All Seven Subsets)}
\label{tab:ood_full}
Table~\ref{tab:ood_complete} provides the complete OOD evaluation including all seven subsets: four non-human species, two human cell lines with divergent PTM loads (reference line 0\% PTM vs HeLa cancer 36\% PTM), and two \textit{E.\ coli} sources with different PTM levels (0\% vs 25.5\%).

\begin{table}[h]
    \centering
\caption{\crchange{Complete heterogeneous zero-shot domain-shift results (all seven OOD subsets). H.\ sapiens (Ref)=reference cell line (0\% PTM); E.\ coli (I)=PXD014877 (0\% PTM); E.\ coli (II)=PXD027742 (25.5\% PTM); HeLa=PXD027742 (36\% PTM).}}
    \label{tab:ood_complete}
    \tiny
    \setlength{\tabcolsep}{1pt}
    \resizebox{\textwidth}{!}{%
    \begin{tabular*}{\textwidth}{@{\extracolsep{\fill}}lcccccccc@{}}
        \toprule
        \multicolumn{9}{c}{\textbf{MassIVE-KB Trained}} \\
        \midrule
        \textbf{Model} & \textbf{ID} & \textbf{H.\ sap. (Ref)} & \textbf{E.\ coli (I)} & \textbf{C.\ elegans} & \textbf{A.\ thaliana} & \textbf{Yeast} & \textbf{E.\ coli (II)} & \textbf{HeLa} \\
        \midrule
        Prosit & 0.100 & 0.465/0.535/+0.029 & 0.474/0.526/$-$0.001 & 0.471/0.529/+0.006 & 0.484/0.516/$-$0.035 & 0.477/0.523/$-$0.020 & 0.477/0.523/$-$0.011 & 0.471/0.529/+0.006 \\
        Prosit T. & \textbf{0.098} & 0.466/0.534/+0.024 & 0.477/0.523/$-$0.005 & 0.472/0.528/+0.004 & 0.485/0.515/$-$0.038 & 0.481/0.519/$-$0.029 & 0.482/0.518/$-$0.021 & 0.475/0.525/$-$0.006 \\
        PredFull & 0.167 & \textbf{0.147}/\textbf{0.853}/\textbf{0.869} & \textbf{0.184}/\textbf{0.816}/\textbf{0.794} & \textbf{0.222}/\textbf{0.778}/\textbf{0.714} & \textbf{0.203}/\textbf{0.797}/\textbf{0.751} & \textbf{0.281}/\textbf{0.719}/\textbf{0.554} & \textbf{0.292}/\textbf{0.708}/\textbf{0.527} & \textbf{0.220}/\textbf{0.780}/\textbf{0.717} \\
        AlphaPep. & 0.129 & 0.462/0.538/+0.042 & 0.471/0.529/+0.011 & 0.470/0.530/+0.019 & 0.482/0.518/$-$0.026 & 0.473/0.527/$-$0.006 & 0.475/0.525/$-$0.004 & 0.468/0.532/+0.017 \\
        UniSpec & 0.337 & 0.383/0.618/+0.062 & 0.379/0.621/+0.066 & 0.382/0.618/+0.017 & 0.384/0.616/$-$0.016 & 0.393/0.607/+0.018 & 0.397/0.603/+0.052 & 0.380/0.620/+0.097 \\
        FastSpel & 0.476 & 0.466/0.534/$-$0.005 & 0.463/0.537/$-$0.001 & 0.476/0.524/$-$0.037 & 0.469/0.531/$-$0.034 & 0.461/0.539/$-$0.007 & 0.466/0.534/$-$0.016 & 0.461/0.539/+0.007 \\
        \midrule
        \multicolumn{9}{c}{\textbf{PROSPECT Trained}} \\
        \midrule
        \textbf{Model} & \textbf{ID} & \textbf{H.\ sap. (Ref)} & \textbf{E.\ coli (I)} & \textbf{C.\ elegans} & \textbf{A.\ thaliana} & \textbf{Yeast} & \textbf{E.\ coli (II)} & \textbf{HeLa} \\
        \midrule
        Prosit & \textbf{0.138} & 0.180/0.820/+0.821 & 0.282/0.718/+0.565 & \textbf{0.160}/\textbf{0.840}/0.856 & \textbf{0.094}/\textbf{0.906}/\textbf{0.950} & 0.284/0.716/+0.557 & 0.291/0.709/+0.548 & 0.244/0.756/+0.664 \\
        Prosit T. & 0.142 & 0.196/0.804/+0.786 & 0.301/0.699/+0.513 & 0.169/0.831/+0.839 & 0.102/0.898/+0.941 & 0.288/0.712/+0.543 & 0.296/0.704/+0.531 & 0.259/0.741/+0.624 \\
        PredFull & 0.203 & \textbf{0.174}/\textbf{0.826}/\textbf{0.819} & \textbf{0.222}/\textbf{0.778}/\textbf{0.699} & 0.199/0.801/+0.763 & 0.163/0.837/+0.837 & \textbf{0.236}/\textbf{0.764}/\textbf{0.674} & \textbf{0.250}/\textbf{0.750}/\textbf{0.642} & \textbf{0.196}/\textbf{0.804}/\textbf{0.770} \\
        AlphaPep. & 0.278 & 0.241/0.759/+0.688 & 0.309/0.691/+0.491 & 0.246/0.754/+0.668 & 0.157/0.843/+0.862 & 0.319/0.681/+0.462 & 0.315/0.685/+0.481 & 0.274/0.726/+0.589 \\
        UniSpec & 0.193 & 0.196/0.804/+0.787 & 0.303/0.697/+0.512 & 0.188/0.812/+0.805 & 0.120/0.880/+0.919 & 0.294/0.706/+0.533 & 0.294/0.706/+0.541 & 0.251/0.749/+0.648 \\
        FastSpel & 0.379 & 0.362/0.638/+0.072 & 0.352/0.648/+0.034 & 0.392/0.609/+0.022 & 0.344/0.656/+0.006 & 0.370/0.629/+0.024 & 0.385/0.615/+0.018 & 0.361/0.639/+0.069 \\
        \bottomrule
    \end{tabular*}}
\end{table}

\subsection{Covariate-Shift Decomposition of Apparent OOD Advantage}
\label{app:ood_covariate_shift}

The PROSPECT-trained Prosit model achieves a lower median SA on \textit{A.\ thaliana} (0.094) than on its in-domain PROSPECT test set (0.138), seemingly indicating better OOD than ID performance. This subsection demonstrates that the apparent advantage is entirely explained by covariate distribution shift between the two test populations---a Simpson's paradox---and that after matching on key covariates, the expected ID${}>$OOD ordering is restored.

\paragraph{Covariate distribution mismatch.}
Table~\ref{tab:ood_covariate_shift} contrasts the covariate profiles of the PROSPECT ID test set and the \textit{A.\ thaliana} OOD set. The two populations differ substantially along three axes: (1)~NCE: PROSPECT spans six NCE values (20--45) with 52\% at NCE${}=28$ or 35, whereas \textit{A.\ thaliana} is entirely at NCE${}=25$; (2)~PTM: PROSPECT is 50\% PTM, \textit{A.\ thaliana} is 0\%; (3)~Instrument/fragmentation: NCE${}=28$ is 68\% ITMS and NCE${}=35$ is 63\% CID+ITMS in PROSPECT, while \textit{A.\ thaliana} is 100\% Orbitrap/HCD.

\begin{table}[h]
    \centering
    \caption{Covariate distribution comparison between PROSPECT ID test and \textit{A.\ thaliana} OOD.}
    \label{tab:ood_covariate_shift}
    \small
    \setlength{\tabcolsep}{3pt}
    \begin{tabular}{lcc}
        \toprule
        \textbf{Covariate} & \textbf{PROSPECT ID} & \textbf{\textit{A.\ thaliana} OOD} \\
        \midrule
        NCE${}=25$ fraction & 13.5\% & 100\% \\
        NCE${}=28$ fraction & 21.2\% & 0\% \\
        NCE${}=35$ fraction & 31.0\% & 0\% \\
        has\_ptm fraction & 50.0\% & 0.0\% \\
        ITMS fraction & 34.1\% & 0.0\% \\
        CID fraction & 19.6\% & 0.0\% \\
        Mean peptide length & 12.7 & 15.4 \\
        \bottomrule
    \end{tabular}
\end{table}

\paragraph{NCE--instrument confound.}
The poor performance at NCE${}=28$ and 35 is not an artifact of collision energy alone. In PROSPECT, these NCE values are strongly confounded with instrument type and fragmentation method (Table~\ref{tab:nce_instrument_confound}): NCE${}=28$ spectra are 68\% ITMS-detected (median 217 peaks vs.\ 51 for FTMS), and NCE${}=35$ spectra are 63\% CID-fragmentated and ITMS-detected. Both ITMS detection and CID fragmentation produce noisier, more complex spectra that are harder for any model to predict. This confound explains why all six models---not just Prosit---show degraded performance at these NCE values.

\begin{table}[h]
    \centering
    \caption{NCE $\times$ instrument/fragmentation confound in PROSPECT test set. Peak counts reflect spectrum complexity.}
    \label{tab:nce_instrument_confound}
    \small
    \setlength{\tabcolsep}{3pt}
    \begin{tabular}{cccccc}
        \toprule
        \textbf{NCE} & \textbf{$n$} & \textbf{ITMS\%} & \textbf{CID\%} & \textbf{Med.\ peaks} & \textbf{Prosit SA} \\
        \midrule
        20 & 5{,}019 & 0 & 0 & 41 & 0.083 \\
        23 & 5{,}472 & 0 & 0 & 48 & 0.078 \\
        25 & 6{,}723 & 0 & 0 & 51 & 0.077 \\
        28 & 10{,}607 & 68 & 0 & 217 & 0.329 \\
        30 & 6{,}639 & 0 & 0 & 57 & 0.083 \\
        35 & 15{,}476 & 63 & 63 & 134 & 0.332 \\
        \bottomrule
    \end{tabular}
\end{table}

\paragraph{Stratified performance restores expected ordering.}
Table~\ref{tab:ood_stratified_sa} reports PROSPECT ID median SA stratified by NCE and PTM. The ``hard'' subpopulation (NCE${}=28$/35) constitutes 52\% of the test set with SA${\approx}0.33$, inflating the overall median to 0.138. Restricting to the covariate profile matching \textit{A.\ thaliana} (NCE${}=25$, no PTM) yields SA${}=0.070$, which is lower (better) than the \textit{A.\ thaliana} OOD value of 0.094 by 0.024 SA units---consistent with normal OOD degradation. This confirms that the apparent ``OOD${}>$ID'' paradox is a Simpson's paradox induced by distribution shift in NCE, PTM, and instrument type.

\begin{table}[h]
    \centering
    \caption{Prosit PROSPECT ID performance stratified by covariates (median SA).}
    \label{tab:ood_stratified_sa}
    \small
    \setlength{\tabcolsep}{3pt}
    \begin{tabular}{lrc}
        \toprule
        \textbf{Stratum} & \textbf{$n$} & \textbf{Med.\ SA} \\
        \midrule
        Overall & 50{,}000 & 0.138 \\
        NCE${}=25$ only & 6{,}723 & 0.077 \\
        NCE${}=25$, no PTM & 3{,}309 & 0.070 \\
        NCE${}=25$, no PTM, $z{=}2$ & 2{,}248 & 0.058 \\
        NCE${}=28$/35 (hard) & 26{,}083 & 0.330 \\
        \midrule
        \textit{A.\ thaliana} OOD (reference) & 20{,}000 & 0.094 \\
        \bottomrule
    \end{tabular}
\end{table}

\paragraph{Cross-model consistency.}
The NCE${}=28$/35 degradation is not Prosit-specific. Table~\ref{tab:ood_cross_model_nce} shows that all neural models perform substantially worse on the NCE${}=28$/35 subpopulation than on the remaining NCE values, confirming that the confound is a property of the data, not of any particular architecture.

\begin{table}[h]
    \centering
    \caption{Cross-model median SA on PROSPECT ID: hard (NCE${}=28$/35) vs.\ easy (other NCE) subpopulations.}
    \label{tab:ood_cross_model_nce}
    \small
    \setlength{\tabcolsep}{3pt}
    \begin{tabular}{lccc}
        \toprule
        \textbf{Model} & \textbf{Overall} & \textbf{Hard (NCE 28/35)} & \textbf{Easy (other NCE)} \\
        \midrule
        Prosit          & 0.138 & 0.330 & 0.080 \\
        Prosit Trans.   & 0.142 & 0.329 & 0.086 \\
        PredFull        & 0.204 & 0.248 & 0.162 \\
        AlphaPeptDeep   & 0.278 & 0.386 & 0.159 \\
        \bottomrule
    \end{tabular}
\end{table}

\section{\crchange{Target-Space Sensitivity and Projection Scope}}
\label{app:target_space_sensitivity}

\crchange{\crdehedge{The canonical leaderboard scores a shared $b/y$ fragment-intensity subtask.} To quantify how target definition affects conclusions, we evaluate the same 50,000 held-out PROSPECT spectra in three representations: canonical $b/y$, an expanded ion space, and a dense raw-20k space. \crdehedge{The expanded and raw-20k columns show how the ordering moves when more of each model's native output is visible.}}

\begin{table}[h]
    \centering
    \caption{\crchange{Three-space sensitivity diagnostic on the PROSPECT held-out set. Values are median SA (lower is better); canonical and raw-20k confidence intervals use 2,000 bootstrap resamples where per-spectrum outputs were available.}}
    \label{tab:three_space_sensitivity}
    \scriptsize
    \setlength{\tabcolsep}{2.5pt}
    \resizebox{\columnwidth}{!}{%

    \begin{tabular}{lccc r}
        \toprule
        \textbf{Model} & \textbf{Canonical $b/y$ SA} & \textbf{Expanded-ion SA} & \textbf{Raw-20k SA} & \textbf{$N$} \\
        \midrule
        Prosit & 0.1387 [0.1367,0.1407] & 0.2470 [0.2453,0.2483] & 0.3411 [0.3392,0.3433] & 50,000 \\
        Prosit Transformer & 0.1463 [0.1444,0.1486] & 0.2461 [0.2448,0.2477] & 0.3401 [0.3382,0.3423] & 50,000 \\
        AlphaPeptDeep & 0.2779 [0.2758,0.2800] & 0.4002$^{a}$ & 0.2376 [0.2356,0.2394] & 50,000 \\
        PredFull & 0.2035 [0.2022,0.2049] & 0.2264 [0.2252,0.2276] & 0.3081 [0.3068,0.3095] & 50,000 \\
        FastSpel & 0.3792 [0.3784,0.3800] & 0.4032$^{a}$ & 0.4390 [0.4383,0.4397] & 47,276 \\
        UniSpec & 0.1933$^{b}$ & 0.3864$^{a}$ & 0.4355 [0.4308,0.4408] & 50,000 \\
        \bottomrule
    \end{tabular}}
    \\[-2pt]
    \parbox{\columnwidth}{\scriptsize $^{a}$Separate verified inference run without per-spectrum bootstrap CI. $^{b}$Canonical value from the finalized benchmark run; the separate three-space UniSpec inference was stochastic and is not used to replace the benchmark value.}
\end{table}

\begin{table}[h]
    \centering
    \caption{\crchange{PredFull intensity retained by the canonical projection, computed over 50,000 held-out spectra.}}
    \label{tab:projection_retention}
    \small

    \begin{tabular}{lcc}
        \toprule
        \textbf{Metric} & \textbf{PROSPECT} & \textbf{MassIVE-KB} \\
        \midrule
        Predicted-intensity retention & 0.1984 & 0.1390 \\
        Ground-truth canonical coverage & 0.2965 & 0.1781 \\
        \bottomrule
    \end{tabular}
\end{table}

\crchange{For raw-20k evaluation, experimental spectra are max-normalized, binned at 0.1 Da over 0--2000 $m/z$, and square-root transformed. Sequence-centric predictions are placed at theoretical ion positions and unsupported bins are zero. The ordering changes across target definitions: PredFull is mid-ranked in canonical $b/y$ space but best among the verified models in expanded-ion and raw-20k diagnostics. \crdehedge{The ranking is a property of the target definition as much as of the model.}}

\section{\crchange{Controlled Split and Data-Budget Sensitivity}}
\label{app:split_sensitivity}

\crchange{\crdehedge{The split ablation uses Prosit on PROSPECT. The reported confidence intervals quantify test-sample uncertainty over a single training run.}}

\begin{table}[h]
    \centering
    \caption{\crchange{Two-seed split sensitivity for Prosit/PROSPECT. Median SA is lower-is-better.}}
    \label{tab:split_two_seed}
    \small
    \setlength{\tabcolsep}{3pt}

    \begin{tabular}{lccc}
        \toprule
        \textbf{Split} & \textbf{Seed 42} & \textbf{Seed 17} & \textbf{Seed-42 95\% CI} \\
        \midrule
        Backbone-disjoint & 0.1376 & 0.1391 & [0.1355, 0.1397] \\
        Row-random & 0.1254 & 0.1282 & [0.1239, 0.1272] \\
        Modified-sequence & 0.1392 & 0.1424 & [0.1369, 0.1412] \\
        \bottomrule
    \end{tabular}
\end{table}

\begin{table}[h]
    \centering
    \caption{\crchange{Train--test overlap audit for the completed PROSPECT split construction.}}
    \label{tab:split_overlap_audit}
    \small
    \setlength{\tabcolsep}{3pt}

    \begin{tabular}{lccc}
        \toprule
        \textbf{Split} & \textbf{Backbone} & \textbf{Modified sequence} & \textbf{Sequence+charge+NCE key} \\
        \midrule
        Row-random & 61.67\% & 57.85\% & 17.84\% \\
        Modified-sequence & 10.37\% & 0.00\% & 0.00\% \\
        Backbone-disjoint & 0.00\% & 0.00\% & 0.00\% \\
        \bottomrule
    \end{tabular}
\end{table}

\begin{table}[h]
    \centering
    \caption{\crchange{Stricter split sensitivity for Prosit/PROSPECT at a 100k training-row budget.}}
    \label{tab:stricter_split_sensitivity}
    \small

    \begin{tabular}{lrr}
        \toprule
        \textbf{Condition} & \textbf{Unique backbones} & \textbf{Median SA} \\
        \midrule
        Backbone control & 77,663 & 0.1688 \\
        AA-composition split & 70,371 & 0.1683 \\
        6-mer-disjoint split & 7,602 & 0.2138 \\
        Backbone-count-matched low-data control & 7,602 & 0.2616 \\
        \bottomrule
    \end{tabular}
\end{table}

\crchange{\crdehedge{The 6-mer result is a stress-test variant of the primary protocol. Its approximately ten-fold lower backbone diversity, together with the lower-data matched control, sets how far the ordering can move at this extreme; we report it as a bound on the trend.}}

\begin{table}[h]
    \centering
    \caption{\crchange{Budget sensitivity on the same PROSPECT held-out test set. Results use one training seed (42).}}
    \label{tab:budget_sensitivity}
    \small

    \begin{tabular}{lcc}
        \toprule
        \textbf{Training budget} & \textbf{Prosit SA} & \textbf{Prosit Transformer SA} \\
        \midrule
        50k & 0.1958 & 0.1879 \\
        100k & 0.1688 & 0.1623 \\
        500k & 0.1376 & 0.1419 \\
        \bottomrule
    \end{tabular}
\end{table}

\crchange{The relative ordering changes across fixed budgets: Prosit Transformer is better at 50k and 100k, whereas Prosit is better at 500k. \crdehedge{These three budgets are the range we verified.}}

\section{\crchange{Community Contribution Guide}}
\label{app:contribution_guide}

\crchange{New models can be contributed through an output adapter that maps native predictions into the shared evaluator while retaining native-space diagnostics. New datasets should provide the benchmark parquet schema and follow the same global filtering, PTM normalization, split construction, and evaluator interface. This design allows future methods and data sources to be compared under the existing canonical task while preserving broader native-output analyses.}

\section{Projection Details}
\label{app:projection_details}

Table~\ref{tab:projection_summary} summarizes the model-specific projection $\Pi_m$ for each baseline. All projections share the same output: a 234-dimensional vector in the canonical space $\mathcal{C}$.

\begin{table}[h]
\centering
\caption{Per-model projection from native output to the shared canonical space $\mathcal{C}$ ($\dim = 234$).}
\label{tab:projection_summary}
\small
\begin{tabular}{@{}lll@{}}
\toprule
\textbf{Model} & \textbf{Native Output} & \textbf{Projection $\Pi_m$} \\
\midrule
Prosit / Prosit Trans. & 174-d ion tensor ($L{\le}30$) & Extend to 234-d (extend to $L{=}40$) \\
AlphaPeptDeep & Dynamic-length ion tensor & Truncate/pad to 234-d \\
PredFull & 20,000-bin full spectrum & Extract $b/y$ bins by theoretical $m/z$ \\
UniSpec & Dictionary-indexed vector & Map dictionary entries to canonical positions \\
FastSpel & Bucket-based ion vector & Align to canonical layout, zero-fill gaps \\
\bottomrule
\end{tabular}
\end{table}

\noindent\textbf{Ground-truth projection.} Experimental spectra are first binned into a uniform $m/z$ grid (0.1\,Da bins, 0--2000\,Da), then intensities at theoretical $b/y$ ion positions are extracted to form $\mathbf{I}^{(c)} \in \mathcal{C}$. This model-agnostic projection ensures that ground truth is treated identically regardless of which model is being evaluated.

\noindent\textbf{Valid positional mask.} A mask $\mathcal{M}(\mathbf{x})$ excludes physically impossible positions: cleavage site $p > L(s) - 1$ or fragment charge $z_f > z$. All metrics are computed exclusively over valid positions.

\section{Adaptation of Baseline Models}
\label{app:model_adaptation}

\subsection{PredFull: Global Mass Conditioning}
\label{app:predfull}
\paragraph{Original formulation.}
PredFull is a CNN-based full-spectrum predictor whose native target is a dense 20,000-bin intensity vector over the $m/z$ range 0--2000 Da.

\paragraph{Benchmark adaptation.}
The original implementation was tied to rigid metadata handling and limited PTM support. To support the benchmark PTM scope without redesigning the convolutional backbone, we introduced a global mass-conditioning strategy: N-terminal acetylation is reflected through precursor-mass correction and an auxiliary meta-feature flag, while the data loader is modified to ingest normalized collision energy directly from the benchmark tables.

\paragraph{Output used in PepSpecBench.}
PredFull is trained in its native full-spectrum format. For evaluation, we extract the subset of bins corresponding to the theoretical $b/y$ ions and project them to the shared canonical space $\mathcal{C}$; all reported metrics use this projected output.

\paragraph{Full-spectrum binning details.}
The 20,000-bin representation uses uniform 0.1\,Da bins spanning $m/z$ 0--2000\,Da. To project canonical $b/y$ ions from the full-spectrum output, we compute theoretical $m/z$ for each ion and map to the nearest bin index via $\mathrm{round}(m_z / 0.1)$. This 0.1\,Da resolution provides sufficient precision for the benchmark's canonical ion extraction while maintaining computational efficiency. The same binning configuration is applied identically to all full-spectrum models (PredFull) and to ground-truth spectra during canonical projection, ensuring consistent treatment across model outputs and reference data.

\subsection{Prosit: Extended Canonical Ion Tensor}
\label{app:prosit}
\paragraph{Original formulation.}
Prosit predicts relative intensities of canonical $b/y$ ions and is classically associated with a fixed-size output for peptides up to length 30 and fragment charges up to 3.

\paragraph{Benchmark adaptation.}
PepSpecBench extends the peptide-length scope to 40 residues. We therefore replace the original fixed-length target with an extended canonical ion-tensor target that remains restricted to $b/y$ ions and fragment charges up to 3, but now covers all cleavage positions up to length 40. Sequence loading and metadata handling are standardized to the benchmark parquet interface.

\paragraph{Output used in PepSpecBench.}
Prosit is trained with an extended canonical ion tensor (covering peptide lengths up to 40) and projected to the shared canonical space $\mathcal{C}$; all reported metrics use this projected output.

\subsection{AlphaPeptDeep: Flexible Output Adapter}
\label{app:alphapeptdeep}
\paragraph{Original formulation.}
AlphaPeptDeep provides a flexible PTM-aware prediction procedure rather than a single fixed benchmark vector. Its output space is organized around charged fragment channels and can be reshaped according to sequence length and supported ion types.

\paragraph{Benchmark adaptation.}
For PepSpecBench, we standardized its configuration to ensure comparability:
\begin{itemize}
    \item \textbf{Layers:} 4 Transformer layers with 8 heads.
    \item \textbf{Hidden Dimension:} 256.
    \item \textbf{PTM Handling:} Unlike PredFull, AlphaPeptDeep treats PTMs as explicit tokens in the sequence. We restricted its vocabulary to the benchmark's standard PTM set (Acetylation, Oxidation, Carbamidomethylation) to prevent it from leveraging obscure modifications not available to other models.
\end{itemize}
\paragraph{Input metadata wiring.}
AlphaPeptDeep receives precursor charge and normalized collision energy (NCE) as explicit metadata inputs alongside the peptide sequence. The benchmark adapter extracts \texttt{precursor\_charge} (or \texttt{charge}) and \texttt{collision\_energy} (or \texttt{nce}) from the parquet tables, normalizes NCE to the [0,100] range, and constructs the precursor DataFrame expected by AlphaPeptDeep's API. This ensures that metadata are correctly passed to the model during both training and inference.

\paragraph{Charge-state mode collapse diagnosis.}
Despite correct input wiring, AlphaPeptDeep exhibits severe charge-state mode collapse (98.0\% high-SAS fraction under $z{=}2{\rightarrow}3$ perturbation, Section~\ref{subsec:model_sensitivity}). This indicates that the model effectively ignores the charge-state input during inference, over-relying on the high-dimensional sequence representation. We confirmed that the charge value is correctly propagated through the data pipeline; the collapse is therefore an architectural/training phenomenon rather than an implementation error. This finding highlights a limitation of flexible sequence-centric architectures: they may fail to properly condition on low-dimensional control variables even when these are explicitly provided as inputs.

\paragraph{Output used in PepSpecBench.}
AlphaPeptDeep outputs are mapped to an intermediate canonical ion tensor and then projected to the shared canonical space $\mathcal{C}$; all reported metrics use this projected output.

\subsection{Prosit Transformer: Extended Canonical Ion Tensor}
\label{app:prosittransformer}
\paragraph{Original formulation.}
Prosit Transformer follows the Prosit-style backbone-ion prediction setup, with a Transformer encoder-decoder replacing the GRU while still targeting canonical $b/y$ ions.

\paragraph{Benchmark adaptation.}
We adapted the implementation to support our standardized PTM tokens. While the original model had limited PTM support and shorter sequence constraints, we extended its embedding layer and prediction head to accommodate the benchmark's ``Big 3'' modifications and the expanded length range up to 40 residues.

\paragraph{Output used in PepSpecBench.}
Like Prosit, Prosit Transformer is trained with an extended canonical ion tensor (peptide length up to 40) and projected to the shared canonical space $\mathcal{C}$; all reported metrics use this projected output.

\subsection{UniSpec: Dictionary Reconstruction and Source-Specific Output}
\label{app:unispec}
\paragraph{Original formulation.}
UniSpec predicts spectral intensities through a learned ion dictionary rather than a fixed canonical backbone tensor. In its original form, this dictionary can cover a broad set of ions, including non-canonical peaks and loss patterns.

\paragraph{Benchmark adaptation.}
Directly applying the pre-trained model would create dictionary gaps on the benchmark data, where theoretically relevant ions are absent from the learned vocabulary. We therefore introduce a dedicated dictionary-reconstruction procedure that rebuilds the output vocabulary from benchmark-compliant training data and aligns energy conditioning with the benchmark metadata format.

\paragraph{Output used in PepSpecBench.}
All UniSpec outputs are projected to the shared canonical space $\mathcal{C}$ before metric computation, consistent with all other models in the benchmark.

\subsection{FastSpel: Universal Scale Adaptation and Coverage Limitation}
\label{app:fastspel}
\paragraph{Original formulation.}
FastSpel is a lightweight linear baseline that predicts canonical backbone-ion intensities through discrete $(L, z, \mathrm{CE})$ buckets.

\paragraph{Benchmark adaptation.}
To extend support toward the PepSpecBench scope, we removed its internal hard-coded constraints (peptides up to 40 amino acids, precursor charges up to +6) and introduced a CE-Snap matching algorithm during evaluation: collision energy is mapped to the nearest learned discrete bucket, and length/charge are likewise snapped to the closest trained $(L,z)$ combination when needed.

\paragraph{Output used in PepSpecBench.}
FastSpel output is aligned to the shared canonical space $\mathcal{C}$; positions outside the trained bucket combinations are filled with zero. All reported metrics use this projection.

\section{Implementation Details}
\label{app:implementation}

All models were trained on a cluster of NVIDIA A100 (80GB) GPUs. We used standardized training loops with architecture-aware hyperparameters following each model's original publication to ensure fair comparison.

\begin{itemize}
    \item \textbf{Optimizer:} AdamW ($\beta_1=0.9$, $\beta_2=0.999$, weight decay $1\times10^{-2}$) for all neural models.
    \item \textbf{Batch Size:} 1024 (scaled based on memory constraints).
    \item \textbf{Learning Rate Schedule:} Cosine annealing with 5\% linear warmup. Peak learning rates follow the original publications for each architecture.
    \item \textbf{Early Stopping:} All models were monitored using \textbf{median Spectral Angle (SA)} on the validation set with patience=10 epochs. The checkpoint with the best validation SA was selected for testing.
    \item \textbf{Loss Function:} Each model uses its \emph{native} training objective as defined in the original publication, ensuring fair evaluation on intended optimization trajectories.
\end{itemize}

\subsection{Metric Computation and Reliability}
\label{app:metric_reliability}
\paragraph{Pearson correlation coefficient (PCC) computation.}
PCC is computed using standard \texttt{numpy.corrcoef} over masked valid positions in the shared canonical space (level1, 234-dim). The implementation includes edge-case handling: when either prediction or ground truth has near-zero variance ($\sigma < 10^{-8}$), PCC returns 0 rather than NaN. This ensures robust aggregation across diverse spectra while preserving standard correlation semantics.

\paragraph{Interpretation of negative PCC in OOD settings.}
\crchange{Negative PCC values observed for some models on external OOD subsets (e.g., Prosit on Yeast with PCC $=-0.020$) \crdehedge{reflect anti-correlation in the canonical coordinates} rather than computation artifacts. \crdehedge{These subsets differ in PTM load, sequence composition, NCE, acquisition source, instrument, and fragmentation conditions, and the negative values track that composition.}}

\subsection{Hyperparameter Configuration}
\label{app:hyperparams}
Table~\ref{tab:hyperparams} summarizes the key hyperparameters for each baseline model. All configurations follow the respective original publications to ensure each architecture is evaluated on its intended optimization trajectory.

\begin{table}[h]
    \centering
    \caption{Hyperparameter configuration for baseline models. All neural models use AdamW optimizer with cosine annealing schedule and 5\% warmup.}
    \label{tab:hyperparams}
    \small
    \setlength{\tabcolsep}{4pt}
    \begin{tabular}{lccccc}
        \toprule
        \textbf{Model} & \textbf{Core Arch.} & \textbf{Params} & \textbf{Peak LR} & \textbf{Batch Size} & \textbf{Loss Function} \\
        \midrule
        Prosit          & GRU (Encoder-Decoder) & $\sim$2.5M & $1\times10^{-4}$ & 1024 & Masked SA \\
        Prosit Trans.   & Transformer (BERT) & $\sim$86M & $5\times10^{-5}$ & 1024 & Masked SA \\
        PredFull        & CNN (ResNet) & $\sim$8.2M & $3\times10^{-4}$ & 1024 & Cosine Similarity \\
        AlphaPeptDeep   & Modular Transformer & $\sim$1.8M & $1\times10^{-5}$ & 1024 & L1 (native Adam) \\
        UniSpec         & Transformer & $\sim$12M & $1\times10^{-4}$ & 1024 & Neg. Cosine Sim. \\
        FastSpel        & Linear (CPU) & $\sim$50K & N/A & N/A & Least Squares \\
        \bottomrule
    \end{tabular}
\end{table}

\subsection{Training Time}
\label{app:training_time}
Table~\ref{tab:training_time} reports wall-clock training time for all six models on MassIVE-KB and PROSPECT. FastSpel uses pre-trained linear coefficients and requires only minutes; neural models range from $\sim$4.5\,h (AlphaPeptDeep on MassIVE-KB) to $\sim$14\,h (Prosit Transformer on PROSPECT). $^\dagger$PredFull on PROSPECT reached the cluster wall-clock limit of 50{,}000\,s; the reported value is the time-to-best checkpoint, not the full training completion time.

\begin{table}[h]
    \centering
    \caption{Training time (seconds) for all six models. M-KB = MassIVE-KB; PROS = PROSPECT.}
    \label{tab:training_time}
    \begin{tabular}{lrrrr}
        \toprule
        \textbf{Model} & \textbf{M-KB (s)} & \textbf{M-KB (h)} & \textbf{PROS (s)} & \textbf{PROS (h)} \\
        \midrule
        Prosit         & 33{,}103 & 9.2 & 17{,}057 & 4.7 \\
        Prosit Trans.  & 46{,}792 & 13.0 & 49{,}748 & 13.8 \\
        PredFull       & 42{,}884 & 11.9 & 50{,}000$^\dagger$ & 13.9 \\
        AlphaPeptDeep  & 15{,}927 & 4.4 & 32{,}696 & 9.1 \\
        UniSpec        & 17{,}150 & 4.8 & 47{,}223 & 13.1 \\
        FastSpel       & 184 & 0.05 & 383 & 0.11 \\
        \bottomrule
    \end{tabular}
\end{table}

\subsection{In-Domain Accuracy Uncertainty}
\label{app:id_bootstrap_ci}
We report 95\% bootstrap confidence intervals (1000 resamples over the 50k test spectra) for in-domain median SA. Table~\ref{tab:id_bootstrap_ci} shows that intervals are narrow ($\pm$0.001 or less) due to the large test set size.

\begin{table}[h]
    \centering
    \caption{In-domain median SA with 95\% bootstrap CI (1000 resamples over 50k test spectra).}
    \label{tab:id_bootstrap_ci}
    \begin{tabular}{llrrr}
        \toprule
        \textbf{Source} & \textbf{Model} & \textbf{Med.\ SA} & \textbf{95\% CI (lo)} & \textbf{95\% CI (hi)} \\
        \midrule
        \multirow{6}{*}{MassIVE-KB}
            & Prosit         & 0.0995 & 0.0984 & 0.1006 \\
            & Prosit Trans.  & 0.0977 & 0.0969 & 0.0986 \\
            & PredFull       & 0.1673 & 0.1666 & 0.1680 \\
            & AlphaPeptDeep  & 0.1288 & 0.1277 & 0.1300 \\
            & UniSpec        & 0.3369 & 0.3365 & 0.3374 \\
            & FastSpel       & 0.4757 & 0.4752 & 0.4762 \\
        \midrule
        \multirow{6}{*}{PROSPECT}
            & Prosit         & 0.1376 & 0.1354 & 0.1397 \\
            & Prosit Trans.  & 0.1419 & 0.1396 & 0.1444 \\
            & PredFull       & 0.2035 & 0.2022 & 0.2049 \\
            & AlphaPeptDeep  & 0.2779 & 0.2759 & 0.2803 \\
            & UniSpec        & 0.1933 & 0.1912 & 0.1951 \\
            & FastSpel       & 0.3792 & 0.3783 & 0.3801 \\
        \bottomrule
    \end{tabular}
\end{table}

\subsection{Inference Efficiency}
\label{app:efficiency}
Tables~\ref{tab:efficiency_massive} and~\ref{tab:efficiency_prospect} report inference throughput on the MassIVE-KB and PROSPECT-Mini test sets. Prosit is the fastest among neural models on both datasets. PredFull is slower due to subprocess-based evaluation and full-spectrum output.

\begin{table}[h]
    \centering
    \caption{Inference efficiency on the MassIVE-KB test set.}
    \label{tab:efficiency_massive}
    \begin{tabular}{lrr}
        \toprule
        \textbf{Model} & \textbf{Spec/s} & \textbf{Med.\ ms/spec} \\
        \midrule
        Prosit         & 17{,}300 & 0.06 \\
        Prosit Trans.  & 1{,}077  & 0.92 \\
        PredFull       & $\sim$45 & $\sim$22 \\
        AlphaPeptDeep  & 5{,}104  & 0.20 \\
        UniSpec        & 766      & 1.45 \\
        FastSpel       & 47       & 21.4 \\
        \bottomrule
    \end{tabular}
\end{table}

\begin{table}[h]
    \centering
    \caption{Inference efficiency on the PROSPECT test set.}
    \label{tab:efficiency_prospect}
    \begin{tabular}{lrr}
        \toprule
        \textbf{Model} & \textbf{Spec/s} & \textbf{Med.\ ms/spec} \\
        \midrule
        Prosit         & 20{,}100 & 0.05 \\
        Prosit Trans.  & 1{,}087  & 0.92 \\
        PredFull       & 56      & 18.0 \\
        AlphaPeptDeep  & 1{,}835  & 0.55 \\
        UniSpec        & 404     & 2.67 \\
        FastSpel       & 55      & 18.1 \\
        \bottomrule
    \end{tabular}
\end{table}

\subsection{Stratified Performance by Length and Charge}
\label{app:stratified}
Per-model stratification by peptide length ($L \in [6, 40]$), precursor charge ($z=2$--$4$), and PTM vs.\ unmodified is included here as supplementary-only analysis. These summaries are derived from per-sample exports and are intended to complement, not replace, the main-text benchmark tables.

We report stratification results from per-sample export for MassIVE-KB (6 models) and PROSPECT. Tables~\ref{tab:massive_stratified} and~\ref{tab:prospect_stratified} summarize median SA in the shared canonical space by charge and PTM status. On MassIVE-KB (human tryptic), Prosit and Prosit Transformer achieve Med.\ SA 0.072--0.075 at $z=2$, 0.120--0.127 at $z=3$, and 0.150--0.168 at $z=4$, with a small Unmod/PTM gap (0.096--0.098 vs.\ 0.102--0.103). On PROSPECT (balanced PTM coverage), the same models show slightly higher Med.\ SA ($z=2$: 0.116--0.125; $z=3$: 0.159--0.163; $z=4$: 0.178--0.186) and a modest PTM gap (Unmod 0.125--0.129 vs.\ PTM 0.151--0.155). Performance degrades with peptide length on both datasets; on PROSPECT, Prosit Med.\ SA increases from $0.10$ (short peptides) to $0.21$ (long peptides, $L>25$), while on MassIVE-KB the same range spans $0.06$ to $0.18$, consistent with longer peptides presenting a harder prediction target. The same pattern is mirrored by median PCC: on PROSPECT, Prosit drops from 0.9367 to 0.7715 and PredFull from 0.8333 to 0.6152 from short to long peptides, while Prosit Transformer degrades more slowly (0.9189 to 0.8237).

\begin{table}[t]
    \centering
    \caption{MassIVE-KB median SA in the shared canonical space by charge and PTM status.}
    \label{tab:massive_stratified}
    \begin{tabular}{lcccc}
        \toprule
        \textbf{Model} & $z=2$ & $z=3$ & $z=4$ & Unmod / PTM \\
        \midrule
        Prosit         & 0.072 & 0.127 & 0.168 & 0.098 / 0.103 \\
        Prosit Trans.  & 0.075 & 0.120 & 0.150 & 0.096 / 0.102 \\
        PredFull       & 0.144 & 0.189 & 0.226 & 0.164 / 0.176 \\
        AlphaPeptDeep  & 0.096 & 0.162 & 0.217 & 0.128 / 0.132 \\
        UniSpec        & 0.333 & 0.340 & 0.347 & 0.339 / 0.333 \\
        FastSpel       & 0.469 & 0.483 & 0.485 & 0.476 / 0.475 \\
        \bottomrule
    \end{tabular}
\end{table}

\begin{table}[t]
    \centering
    \caption{PROSPECT median SA in the shared canonical space by charge and PTM status. FastSpel is omitted as its discrete bucket architecture yields less informative charge/PTM breakdowns.}
    \label{tab:prospect_stratified}
    \begin{tabular}{lcccc}
        \toprule
        \textbf{Model} & $z=2$ & $z=3$ & $z=4$ & Unmod / PTM \\
        \midrule
        Prosit         & 0.116 & 0.159 & 0.186 & 0.125 / 0.151 \\
        Prosit Trans.  & 0.125 & 0.163 & 0.178 & 0.129 / 0.155 \\
        PredFull       & 0.180 & 0.240 & 0.280 & 0.195 / 0.212 \\
        AlphaPeptDeep  & 0.246 & 0.306 & 0.368 & 0.269 / 0.285 \\
        UniSpec        & 0.173 & 0.216 & 0.238 & 0.182 / 0.205 \\
        \bottomrule
    \end{tabular}
\end{table}

\begin{figure}[t]
    \centering
    \includegraphics[width=0.95\columnwidth]{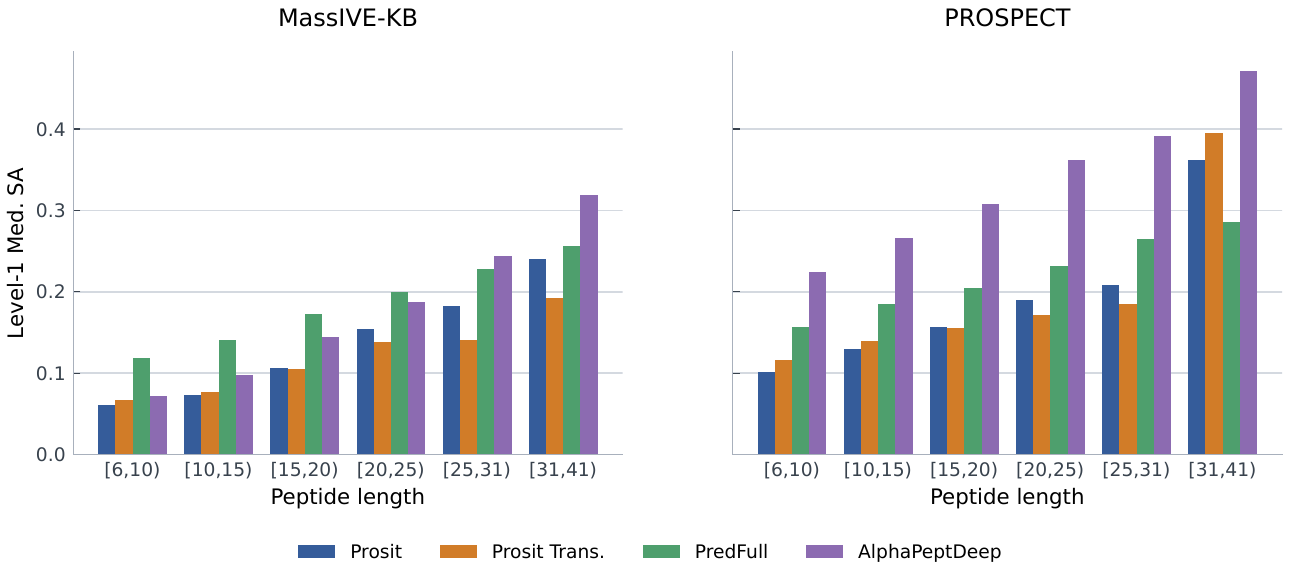}
    \caption{Median SA in the shared canonical space by peptide length bin (MassIVE-KB and PROSPECT). Performance degrades with length; Transformer-based models maintain an advantage across bins.}
    \label{fig:stratification_by_length}
\end{figure}

Figure~\ref{fig:stratification_by_length} visualizes median SA in the shared canonical space by peptide length bin for the top four models on both datasets. Performance degrades monotonically with length: Prosit Med.\ SA increases from $\sim$0.06--0.10 (short peptides) to $\sim$0.18--0.24 (long peptides, $L > 25$) on MassIVE-KB, and from $\sim$0.10--0.12 to $\sim$0.19--0.21 on PROSPECT, consistent with longer peptides presenting a harder prediction target. Transformer-based models (Prosit, Prosit Transformer) maintain an advantage over PredFull and AlphaPeptDeep across length bins.

Figure~\ref{fig:delta_sa_decay} in the main text re-expresses the same degradation as a relative decay rate $\Delta\mathrm{SA} = \mathrm{SA}(L\text{-bin}) - \mathrm{SA}(\text{shortest bin})$, making architecture-specific degradation slopes directly comparable regardless of absolute performance level.

\paragraph{Stratification by PTM type.} Tables~\ref{tab:massive_ptm_stratified} and~\ref{tab:prospect_ptm_stratified} report median SA in the shared canonical space by modification type (Acetylation, Carbamidomethylation, Oxidation, Unmodified). On MassIVE-KB, Prosit and Prosit Transformer show similar performance across PTM types (Med.\ SA 0.095--0.166); Acetylation is slightly harder. On PROSPECT, AlphaPeptDeep exhibits larger gaps (CAM 0.297 vs.\ Unmod 0.269), while Prosit maintains more uniform performance across PTM types.

\begin{table}[t]
    \centering
    \caption{MassIVE-KB median SA in the shared canonical space by PTM type (top-4 models; UniSpec and FastSpel omitted for conciseness).}
    \label{tab:massive_ptm_stratified}
    \begin{tabular}{lcccc}
        \toprule
        \textbf{Model} & \textbf{Ace} & \textbf{CAM} & \textbf{Ox} & \textbf{Unmod} \\
        \midrule
        Prosit         & 0.156 & 0.099 & 0.095 & 0.098 \\
        Prosit Trans.  & 0.166 & 0.095 & 0.106 & 0.096 \\
        PredFull       & 0.199 & 0.176 & 0.170 & 0.164 \\
        AlphaPeptDeep  & 0.187 & 0.127 & 0.128 & 0.128 \\
        \bottomrule
    \end{tabular}
\end{table}

\begin{table}[t]
    \centering
    \caption{PROSPECT median SA in the shared canonical space by PTM type (top-4 models; UniSpec PROSPECT exports and FastSpel PTM breakdown are omitted).}
    \label{tab:prospect_ptm_stratified}
    \begin{tabular}{lcccc}
        \toprule
        \textbf{Model} & \textbf{Ace} & \textbf{CAM} & \textbf{Ox} & \textbf{Unmod} \\
        \midrule
        Prosit         & 0.189 & 0.156 & 0.119 & 0.125 \\
        Prosit Trans.  & 0.198 & 0.150 & 0.125 & 0.129 \\
        PredFull       & 0.200 & 0.230 & 0.211 & 0.195 \\
        AlphaPeptDeep  & 0.296 & 0.297 & 0.262 & 0.269 \\
        \bottomrule
    \end{tabular}
\end{table}

\begin{figure}[t]
    \centering
    \includegraphics[width=0.6\columnwidth]{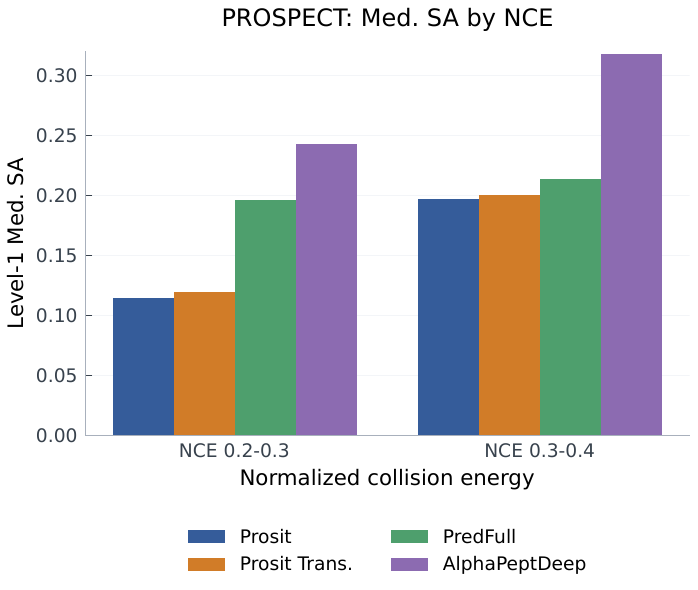}
    \caption{PROSPECT: median SA in the shared canonical space by NCE bin. Performance degrades at higher collision energy.}
    \label{fig:stratification_by_nce}
\end{figure}

\paragraph{Stratification by NCE (PROSPECT).} MassIVE-KB-Mini is effectively single-bin in NCE after reconstruction (all test samples in 0.25--0.30), whereas PROSPECT-Mini spans multiple bins (20.98\% in 0.20--0.25, 34.66\% in 0.25--0.30, and 44.24\% in 0.30--0.35). Figure~\ref{fig:stratification_by_nce} therefore emphasizes PROSPECT for NCE trends. Performance degrades at higher CE under SA and is consistent under PCC: for Prosit, median SA rises from 0.0806 to 0.1974 and median PCC drops from 0.9616 to 0.7888 between 0.20--0.25 and 0.30--0.35.

\subsection{NCE Sensitivity Supplement}
\label{app:nce_sensitivity}

Table~\ref{tab:model_sensitivity_fullrun} provides the numerical summary for the two NCE perturbation experiments introduced in Section~\ref{subsec:model_sensitivity}.

\begin{table}[t]
    \centering
    \caption{Physical-parameter sensitivity. NCE Calibration Analysis: NCE at min SA on PROSPECT (true NCE${}=30$). Blind NCE Perturbation: $\Delta$SA on MassIVE-KB ($25{\rightarrow}30$).}
    \label{tab:model_sensitivity_fullrun}
    \small
    \setlength{\tabcolsep}{3pt}
    \begin{tabular}{@{}lccc@{}}
        \toprule
        \textbf{Model} & \textbf{Calib. min NCE} & \textbf{Calib. SA@30} & \textbf{Blind $\Delta$SA} \\
        \midrule
        Prosit            & 30 & 0.083 & $-$0.000 \\
        Prosit Trans.     & 30 & 0.088 & $+$0.000 \\
        PredFull          & 25 & 0.216 & $+$0.011 \\
        AlphaPeptDeep     & 30 & 0.159 & $+$0.000 \\
        UniSpec           & 35 & 0.325 & $-$0.000 \\
        FastSpel          & 35 & 0.360 & $+$0.000 \\
        \bottomrule
    \end{tabular}
\end{table}

\subsection{Charge Mode Collapse Supplement}
\label{app:charge_mode_collapse}
Table~\ref{tab:charge_mode_collapse_fullrun} provides the full-run summary for the z=2$\rightarrow$3 perturbation experiment introduced in Section~\ref{subsec:model_sensitivity}. The key readout is \texttt{high\_sas\_fraction} with threshold 0.90; higher values indicate stronger charge-insensitive collapse. All four models are evaluated on $N{=}31{,}498$ z=2 spectra, filtered from the PROSPECT-Mini test set (which contains 31{,}498 z=2 entries out of 50{,}000 total). For UniSpec, predictions for all 50{,}000 test spectra were stored in \texttt{vectors\_charge2.npz} together with their \texttt{precursor\_charge} metadata; we filter this file to the $z{=}2$ entries ($N{=}31{,}498$) to match the evaluation scope used for the other three models.

\begin{table}[!t]
    \centering
    \caption{Full-run charge mode collapse summary for the z=2$\rightarrow$3 perturbation. The high-SAS threshold is 0.90, and SAS is computed between two predictions of the same peptide under forced charge metadata change.}
    \label{tab:charge_mode_collapse_fullrun}
    \small
    \setlength{\tabcolsep}{4pt}
    \begin{tabular}{lccccc}
        \toprule
        \textbf{Model} & \textbf{N Eval} & \textbf{Med.\ SAS} & \textbf{Q25} & \textbf{Q75} & \textbf{High-SAS Frac.} \\
        \midrule
        Prosit            & 31,498 & 0.7666 & 0.7129 & 0.8206 & 0.0339 \\
        Prosit Transformer & 31,498 & 0.7483 & 0.6905 & 0.8028 & 0.0220 \\
        AlphaPeptDeep     & 31,498 & 0.9722 & 0.9568 & 0.9834 & 0.9804 \\
     UniSpec & 31,498 & 0.7323 & 0.6666 & 0.7977 & 0.0274 \\
        \bottomrule
    \end{tabular}
\end{table}




\end{document}